\documentclass[10pt,letterpaper]{article}
\usepackage{pslatex}
\usepackage{apacite}
\newcommand{\key}{\textbf}
\usepackage[hidelinks]{hyperref} 
\usepackage{hyperref}       % hyperlinks
\usepackage{url}            % simple URL typesetting
\usepackage{booktabs}       % professional-quality tables
\usepackage{amsfonts}       % blackboard math symbols
\usepackage{nicefrac}       % compact symbols for 1/2, etc.
\usepackage{microtype}      % microtypography
\usepackage{xcolor}         % colors
\usepackage{natbib}
\usepackage{multirow}
\usepackage{graphicx}
\usepackage{floatflt}
\usepackage{wrapfig}
\usepackage{tikz}
\usepackage{mathtools}
\usetikzlibrary{positioning,shapes,arrows}
\usepackage{caption}
\usepackage{subcaption}
\usepackage{physics}
\usepackage{float}

\usepackage{amsmath,amssymb,amsthm}
\usepackage{setspace}
\usepackage{comment}
\usepackage[margin=1.5in]{geometry}
\usepackage{setspace}
\usepackage{cancel} 
\newcommand{\red}{\textcolor{red}}

\newcommand{\blue}{\textcolor{blue}}

\graphicspath{{figures/}}
\usepackage{authblk}
\usepackage[switch]{lineno}
\usepackage{xr}  % External document referencing package

\makeatletter
\newcommand*{\addFileDependency}[1]{%
  \typeout{(#1)}%
  \@addtofilelist{#1}
  \IfFileExists{#1}{}{\typeout{No file #1.}}
}
\makeatother

\newcommand*{\myexternaldocument}[1]{%
  \externaldocument{#1}%
  \addFileDependency{#1.tex}%
  \addFileDependency{#1.aux}%
}

\myexternaldocument{si}

\title{\bf Decomposition of surprisal: Unified computational model of ERP components in language processing}

\author{\large \bf Jiaxuan Li} 
\author{\large \bf Richard Futrell}

\affil{\{jiaxuan.li, rfutrell\}@uci.edu \\ Department of Language Science\\ University of California Irvine}

\author[]{}
\date{}

\begin{document}

\maketitle
\begin{abstract}
The functional interpretation of language-related ERP components has been a central debate in psycholinguistics for decades. We advance an information-theoretic model of human language processing in the brain in which incoming linguistic input is processed at first shallowly and later with more depth, with these two kinds of information processing corresponding to distinct electroencephalographic signatures. Formally, we show that the information content (surprisal) of a word in context can be decomposed into two quantities: (A) \key{shallow surprisal}, which signals shallow processing difficulty for a word, and corresponds with the N400 signal; and (B) \key{deep surprisal}, which reflects the discrepancy between shallow and deep representations, and corresponds to the P600 signal and other late positivities. Both of these quantities can be estimated straightforwardly using modern NLP models. We validate our theory by successfully simulating ERP patterns elicited by a variety of linguistic manipulations in previously-reported experimental data from six experiments, with successful novel qualitative and quantitative predictions. Our theory is compatible with traditional cognitive theories assuming a `good-enough' shallow representation stage, but with a precise information-theoretic formulation. The model provides an information-theoretic model of ERP components grounded on cognitive processes, and brings us closer to a fully-specified neuro-computational model of language processing.
\end{abstract}

%\linenumbers
\section{Introduction}
Human language comprehension is linked to (at least) two distinct and robust event-related potential (ERP) components detectable through electroencephalography: the N400 and P600. The N400 is a negative-going waveform that peaks at around 400 ms after the onset of linguistic signal, whereas the P600 is a positivity at around 600 ms. Since their discovery, a great deal of research has attempted to ascertain the functional interpretation of these signals in order to shed light on the neural mechanisms of human language processing \citep{kutas_event-related_1980,hagoort_syntactic_1993,hoeks_seeing_2004,kim_independence_2005,van_herten_erp_2005,van2012prediction,kuperberg_neural_2007,kuperberg_separate_2016,van2012prediction}. 

In previous work, the N400 and P600 ERP components have been linked to semantic and syntactic anomalies in sentences, and so were taken to indicate some degree of modularity with respect to syntactic and semantic processing~\citep{kutas_event-related_1980,hagoort_syntactic_1993}. However, a number of studies have found P600 effects in response to semantic violations, with or without a corresponding N400~\citep{kim_independence_2005,kuperberg_neural_2007,chow_bag--arguments_2016,brouwer_getting_2012,ryskin2021erp,ito2016predicting,van2012prediction,kuperberg2020tale,delong2020comprehending,brothers2020going}. 
In response to these data, recent psycholinguistic theories have proposed that language comprehension involves a \key{shallow representation}---a plausible representation of the input signal based on a subset of the information in it, which can be formed quickly \citep{van_herten_erp_2005,van_herten_when_2006,kuperberg_separate_2016,brouwer_getting_2012,ferreira_good-enough_2002,ferreira_misinterpretation_2000,ferreira2007good}---along with an error monitoring process \citep{li2023heuristic}.
The shallow representation may reflect comprehenders' inferences about plausible meanings given potentially noisy or errorful input \citep{gibson_rational_2013,ryskin2021erp,li2023heuristic}.
In such theories, the N400 reflects the semantic well-formedness of the shallow representation, whereas P600 reflects a discrepancy between shallow and literal representation. 

However, none of the theories have succeeded in explaining the full range of empirical results \citep[see][for detailed discussion]{brouwer_getting_2012}, and the nature of the shallow representation itself has been under debate and challenging to operationalize.  
Existing studies studies must make subjective decisions about possible candidates for shallow representation~\citep{li2023heuristic,ryskin2021erp}, making it difficult to scale up and account for other studies. In addition, these studies are not integrated with more general computational neuroscientific models of other cognitive processes.%~\citep{ortega2013thermodynamics,futrell2022information,zenon2019information}.

We propose an information-theoretic computational-level model of the N400 and P600 ERP components in language processing, based on a model of language comprehension which proceeds from shallow to deep representation. 
Our model is a generalization of Surprisal Theory, an existing probabilistic model of online language comprehension \citep{hale2001probabilistic,levy2008expectation}. 
We show that surprisal can be decomposed into two parts: (A) the \key{shallow surprisal} of the current word within a shallow representation, predicting the magnitude of the N400 signal, and (B) a \key{deep surprisal} reflecting the difference between the surprisal of the veridical input and the surprisal of the shallow representation, predicting the magnitude of the P600 signal. We run qualitative simulations over previously-reported data from three experiments, featuring semantic and syntactic violations, event structure violations, and semantic relatedness priming. We perform further quantitative analyses on three additional experiments, where we show that our quantities for shallow surprisal and deep surprisal track the N400 and P600 components respectively. 

\section{Model}
\begin{figure}[!htb]
\centering
\includegraphics[width=.75\textwidth]{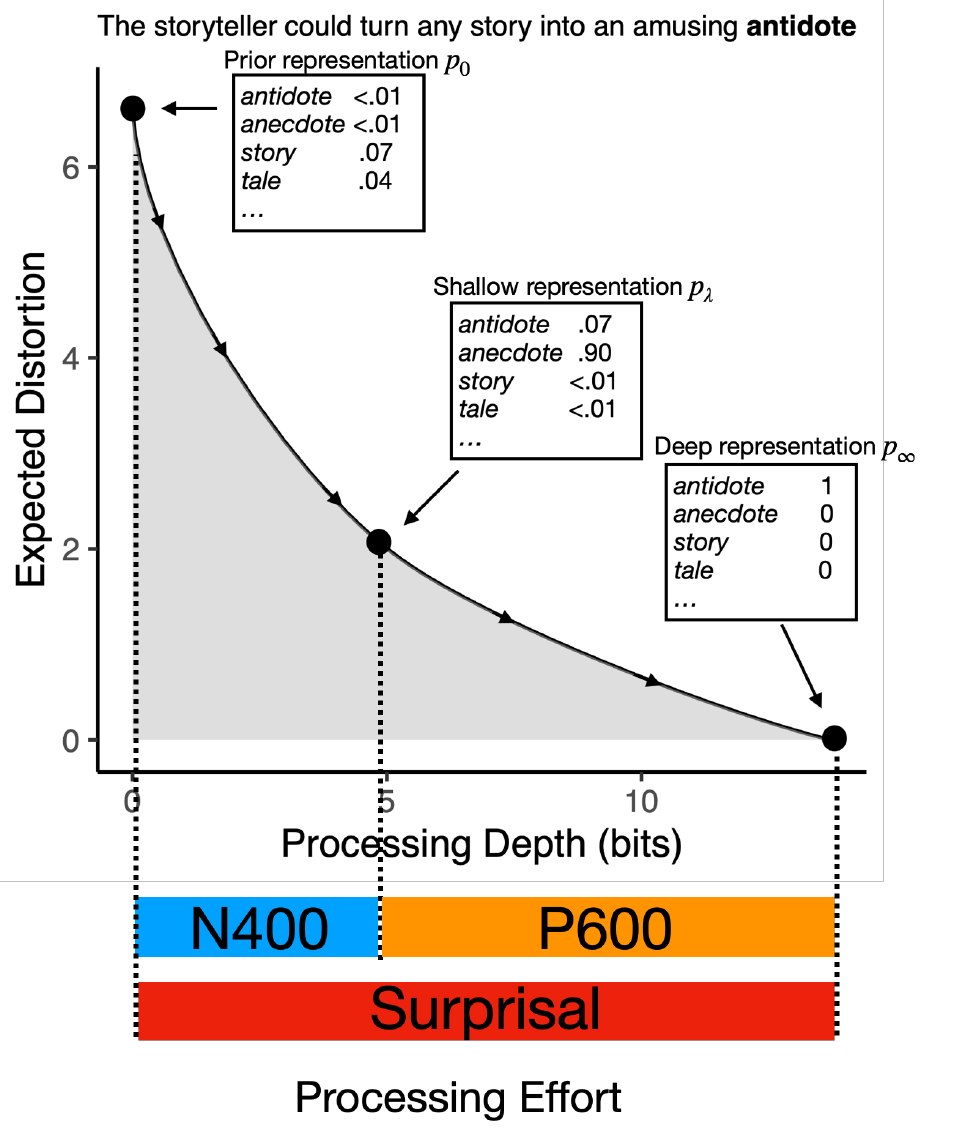}
\caption{Overview of model architecture. The curve represents tradeoff between distortion and processing depth (KL divergence) in optimal representation policies for the given input. Each location in the white part of the plane represents a possible representation policy for the input; the tradeoffs in the gray region are unachievable. The black line shows the \emph{efficient frontier} of policies that achieve the minimal distortion for a given level of processing depth. As a comprehender perceives input $x$, the representation policies move down this frontier, increasing depth and decreasing distortion. The total processing depth is equal to surprisal, and can be partitioned into two parts corresponding to N400 and P600 ERP signals. The figure shows the actual curve and representation policies for the given input, using GPT-2 for the initial representation $p_0$ and a distortion metric based on edit distance.}
\label{fig:overview}
\end{figure}

Consider a comprehender perceiving a sentence, currently observing input word $x$ in the context of (a memory trace of) previously-observed words $c$. For example, in Figure~\ref{fig:overview}, the context is ``The storyteller could turn any story into an amusing \dots'' and the current word is ``antidote''. We propose that the comprehender interprets the input $x$ through a process that extracts as little information from the input as possible while still yielding a reasonably accurate representation in most cases. 

We formalize this idea by characterizing language comprehension using an \key{representation policy}: a probability distribution $p(w \mid c, x)$ on representations $w$ given perceptual inputs $x$ and context $c$. The representation policy is boundedly rational. It minimizes a \textbf{distortion metric}---a measure $d(w,x)$ of how bad it is to form a representation $w$ given input $x$---subject to a constraint on how many bits of information are extracted from the sensory input, corresponding to processing effort. Examples of representation policies and their tradeoffs is shown in Figure~\ref{fig:overview}. Given input $x=$``antidote'', comprehenders may initially perceive ``antidote'' as $w=$``anecdote'', a shallow representation that is consistent with prior expectations but which uses only partial information from the input word.

\paragraph{Form of an optimal representation policy} 
The comprehenders' representation policy $p(w \mid c, x)$ is chosen to minimize expected distortion subject to a constraint on the depth of processing. We model depth of processing as the amount of information extracted about the input $x$ \citep{cheyette2020unified}:
\begin{equation}
\mathrm{Depth} = \mathrm{D}_{\text{KL}}\left[p(w \mid c, x) \| p_0(w \mid c) \right],
\end{equation}
which is the KL divergence \citep{cover2006elements} of the representation policy from the comprehender's prior expectations $p_0(w \mid c)$. For a desired level of average processing depth, the representation policy that minimizes average distortion has the form:
\begin{equation}
    \label{eq:heuristic-policy}
    p_\lambda(w \mid c, x) \propto p_0(w \mid c) e^{-\lambda d(w, x)},
\end{equation}
where the scalar $\lambda$ is selected to achieve the target processing depth.

Our theory defines an infinite family of representation policies corresponding to increasing processing depth. Among these, we define three critical representation policies, shown in Fig.~\ref{fig:overview}, which are crucial for our model of ERPs and which give intuition for model behavior. When $\lambda = 0$ (prior representation), the policy reduces to $p_\lambda(w \mid c, x) = p_0(w \mid c)$, meaning that the comprehender is taking into account no information from the current input. As $\lambda \rightarrow \infty$ (veridical representation), the policy $p_\infty(w \mid x, c)$ concentrates all probability mass on representations with minimum distortion, that is, on veridical representations. When $\lambda$ is moderate (shallow representation), the representation reflects a trade-off between prior expectations and processing depth.

\paragraph{Surprisal decomposition} We hold that our measure of processing depth corresponds to the effort involved in processing linguistic input, as reflected in reading times and ERPs. This is in accord with Surprisal Theory, an influential theory of language processing, which holds that processing effort for a current word $x$ in context $c$ will be proportional to the surprisal of the current word, $-\log p(x \mid c)$ \citep{hale2001probabilistic,levy2008expectation,frank2011insensitivity,smith2013effect,wilcox2020predictive,shain2022robust,wilcox2023testing,xu2023linearity}. This surprisal is the processing depth required to go from prior expectations $p_0$ to the veridical representations $p_\infty$ \citep{levy2008expectation}:
\begin{equation}
    \mathrm{D}_{\text{KL}}\left[p_{\infty} \| p_0\right] = -\ln p(x \mid c).
\end{equation}

We maintain the idea that the total amount of processing effort is given by surprisal, but we partition the effort into two parts, as shown in Figure~\ref{fig:overview}. The two parts correspond to (A) the depth of processing required for the shallow representations, which we call \key{shallow surprisal}:
\begin{equation}
\label{eq:decomposition}
A = \underbrace{ \mathrm{D}_{\text{KL}}\left[p_{\lambda} \| p_0\right]}_{\text{shallow surprisal}}
\end{equation}
and (B) the remaining depth of processing required for the veridical representation, which we call the \key{deep surprisal}:
\begin{equation}
B = \underbrace{ \mathrm{D}_{\text{KL}}\left[p_{\infty} \| p_0\right] - \mathrm{D}_{\text{KL}}\left[p_{\lambda} \| p_0\right]}_{\text{deep surprisal}},
\end{equation}
such that $\mathrm{D}_{\text{KL}}[p_\infty \| p_0] = A + B$. We propose that the N400 magnitude is proportional to the shallow surprisal $A$ and the P600 magnitude is proportional to the deep surprisal $B$ for distinct positive scalars $\alpha$ and $\beta$ in:
\begin{equation}
\label{eq:proportions}
\mathrm{N400} = \alpha A, \mathrm{P600} = \beta B.
\end{equation}

\paragraph{Distortion metric} We use a distortion metric that reflects a mixture of semantic and phonological distance between $w$ and $x$:
\begin{equation}\label{eq:likelihood}
d(w, x) = d_\varphi(w, x) + \gamma d_\sigma(w,x), 
\end{equation}
where $d_\varphi$ is a form-based phonological or orthographic distance metric, $d_\sigma$ is a semantic distance metric, and $\gamma$ is a scalar conversion factor between semantic and form-based distance. We fix $\gamma=8$ in all simulations below.

In addition to operationalizing shallow processing, the representation policy can be seen as performing a kind of fast error correction on the input. If the distortion metric $d(w, x)$ represents the log likelihood that the listener receives input $x$ when the speaker intended to produce $w$, then Eq.~\ref{eq:heuristic-policy} reduces to Bayes' rule at $\lambda=1$, describing rational error correction in a noisy channel \citep{levy_noisy-channel_2008,gibson_rational_2013,norris2008perception,bicknell2010rational,poppels2016structure,li2023heuristic,ryskin2021erp}, as shown in SI~Section~1. Indeed, the two terms in the distortion metric reflect common sources of errors in speech, phonological and semantic interference \citep{dell1981stages}.

\subsection*{Implementation}
We use Levenshtein edit distance for the form-based distance $d_\varphi$, cosine distance between GPT-2 embeddings for semantic distance $d_\sigma$,
and GPT-2 probabilities for the prior $p_0$. We approximate the KL divergence in Eq.~\ref{eq:decomposition} by sampling candidate representations $w$ generated by prompting a large language model and reweighting these candidates using Eq.~\ref{eq:heuristic-policy}. See Materials \& Methods for further details.

\section{Empirical Validation}
\subsection{Dataset}

Our dataset consists of seven ERP studies. The studies cover a wide range of phenomena including word predictability, plausibility, role reversal, word substitution, word exchange within and across relative clause boundaries, morpho-syntactic violation, semantic violation that is orthographically or semantically related to predictable target, and semantic violation with or without discourse context. Table~\ref{tab:stimuli} shows a list of conditions with sample stimuli and empirical ERP patterns across experiments. The ERP effects in the experimental conditions are all calculated in terms of differences to the ERP signal in the control condition. We conducted qualitative analysis on the first three experiments and quantitative analysis for the last one.

The first study (hereby \emph{AD-98})~\citep{ainsworth1998dissociating} includes four conditions: one control condition (\emph{Control}); one condition with violation of semantic content (\emph{Semantic}); one with syntactic manipulation that shows a P600 effect (\emph{Syntactic}); and one with both semantic and syntactic violation (\emph{Double}). The conditions with semantic violation (\emph{Semantic} and \emph{Double}) elicited an N400 effect relative the \emph{Control} condition, whereas the conditions with syntactic violation (\emph{Syntactic} and \emph{Double}) triggered a P600 effect.

In the second study (hereby \emph{Kim-05}), from~\citep{kim_independence_2005}, there are three conditions: \emph{Attractive}, \emph{Non-attractive}, \emph{Control}. In experimental conditions the animacy of the subject is violated. Additionally, the semantic association between subject and target verb is manipulated such that subject and verb could form a plausible event (in the \emph{Attractive} condition) or not (in \emph{Non-attractive}). While the \emph{Attractive} condition elicited a greater P600 response compared to the control condition, the \emph{Non-attractive} condition elicited a greater N400. 

The third study (hereby \emph{Ito-16}) from~\citep{ito2016predicting} includes four conditions. The three experimental conditions all change one target word in the \emph{Control} condition into a semantically implausible one. In \emph{Semantic-related} condition, the semantic violation is semantically related to the target in \emph{Control} condition. In \emph{Form-related} condition, the semantic error shares orthographic form with the \emph{Control} target. In \emph{Unrelated} condition, the violation is not related to the target in the \emph{Control} condition. All three experimental manipulations triggered N400 effects. In addition, the size of N400 effect to semantically related violation was reduced, and form-related violation triggered a P600 effect as well. 

The fourth study (hereby \emph{Brothers-20}) includes three separate experiments. The first experiment (Brothers-20S) investigates how a minimal, \textbf{single} sentence context (\emph{James unlocked the ...}) affects ERP responses to unexpected but congruent continuations, and to anomalous implausible words. The contextual constraint was solely provided by the verb \emph{unlocked}. There was an N400 effect in both \emph{Unexpected} and \emph{Anomalous} conditions. In the second experiment \emph{Brothers-20L}, the single sentence used in \emph{Brothers-20S} was presented after a low constraint two-sentence discourse. In addition to N400 effects reported in \emph{Brothers-20S}, a significant P600 effect was observed in anomalous continuation in a discourse where the constraint is also \textbf{local} provided by a single verb. In the third experiment \emph{Brothers-20G}, the discourse introduces globally constraining, semantically rich contexts. Similar to \emph{Brothers-20S}, a biphasic N400-P600 effect was observed for anomalous targets, whereas unexpected words elicited an N400 effect.  

The fifth study (hereby \emph{Chow-16R}) includes two manipulations: reversal (\emph{Reversal}) that elicited an N400 effect, and word substitution (\emph{Substitution-1}) that elicited a biphasic effect~\citep{chow_bag--arguments_2016}. In Reversal, the argument roles of nouns in sentences of the experimental condition are reversed relative to what would be expected in a canonical situation. In Substitution-1, one of the arguments in the embedded clause was substituted into a less relevant word, which significantly lowers the semantic association between the target verb and the preceding context. 

The sixth study (hereby \emph{Chow-16S}) includes substitution (\emph{Substitution-2}) with a biphasic effect and swap (\emph{Swap}) with an N400 effect. Substitution-2 is the same word substitution manipulation as Substitution-1 but with different set of experimental items. In Swap, one argument in the embedded clause was swapped with the main subject, thus keeping lexical content the same between conditions. 

\begin{figure}[!htb]
     \centering
     \captionsetup[subfigure]{justification=centering}
    \begin{subfigure}[b]{0.15\linewidth}
         \centering
         \includegraphics[width=\linewidth]{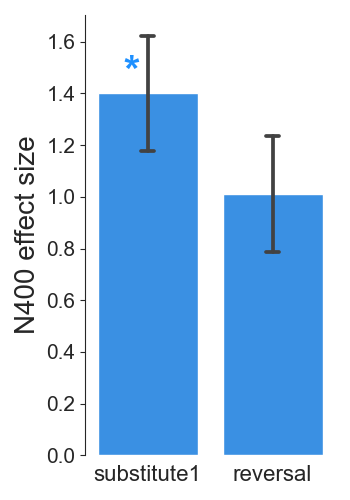}
         \includegraphics[width=\linewidth]{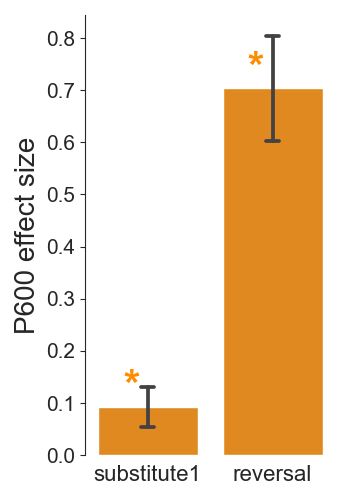}
         \caption{Chow-16R (Model)}
         \label{fig:rev_gpt3}
     \end{subfigure}
     \hfill
     \begin{subfigure}[b]{0.15\linewidth}
         \centering
         \includegraphics[width=\linewidth]{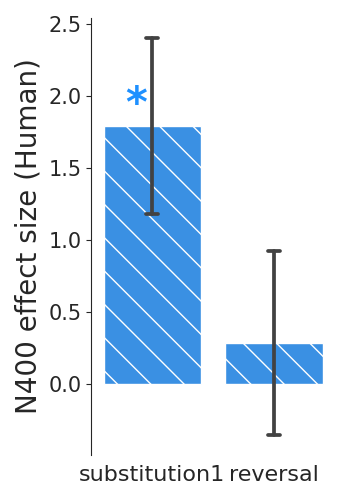}
         \includegraphics[width=\linewidth]{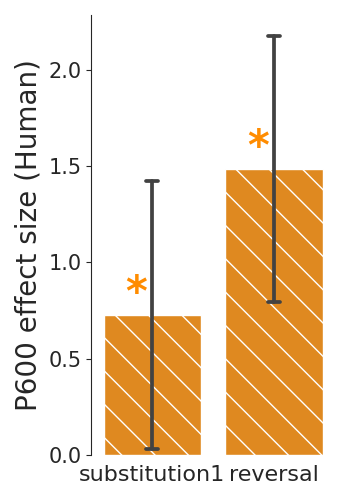}
         \caption{Chow-16R (Human)}
         \label{fig:rev_human}
     \end{subfigure}
     \hfill
    \begin{subfigure}[b]{0.15\linewidth}
         \centering
         \includegraphics[width=\linewidth]{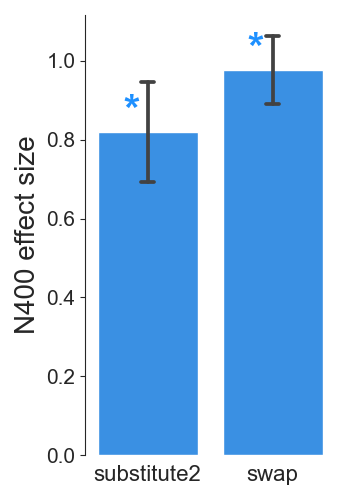}
         \includegraphics[width=\linewidth]{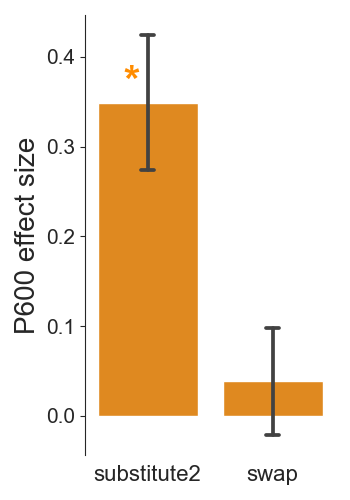}
         \caption{Chow-16S (Model)}
         \label{fig:swap_gpt3}
     \end{subfigure}
     \hfill
     \begin{subfigure}[b]{0.15\linewidth}
         \centering
         \includegraphics[width=\linewidth]{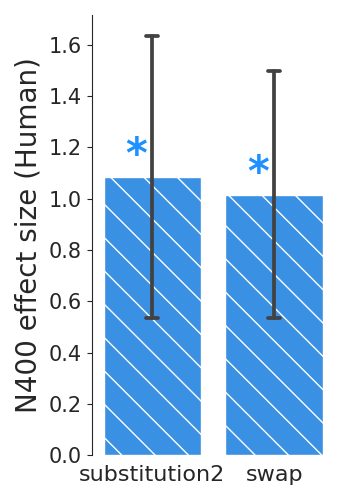}
         \includegraphics[width=\linewidth]{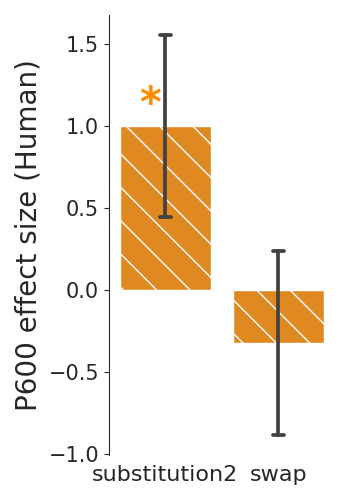}
         \caption{Chow-16S (Human)}
         \label{fig:swap_human}
     \end{subfigure}    
     \vfill
     \begin{subfigure}[b]{0.25\linewidth}
         \centering
         \includegraphics[width=\linewidth]{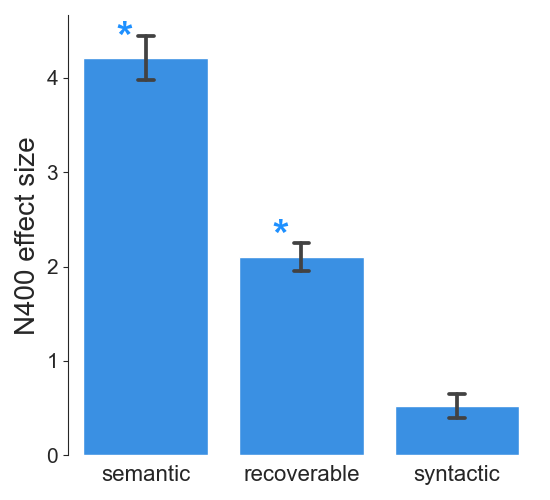}
         \includegraphics[width=\linewidth]{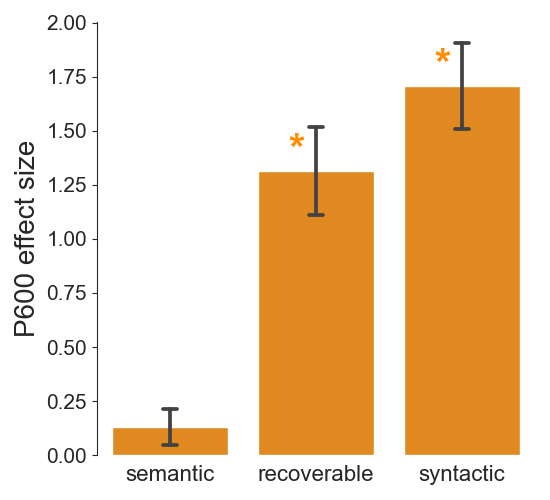}
         \caption{Ryskin-21 (Model)}
         \label{fig:ryskin_gpt3}
     \end{subfigure}
     %\hfill
     \begin{subfigure}[b]{0.25\linewidth}
         \centering
         \includegraphics[width=\linewidth]{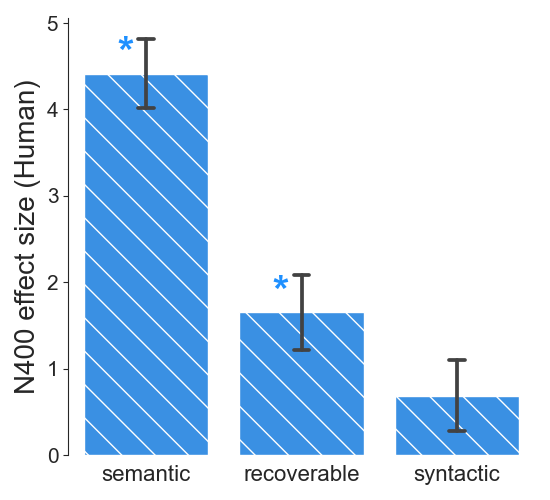}
         \includegraphics[width=\linewidth]{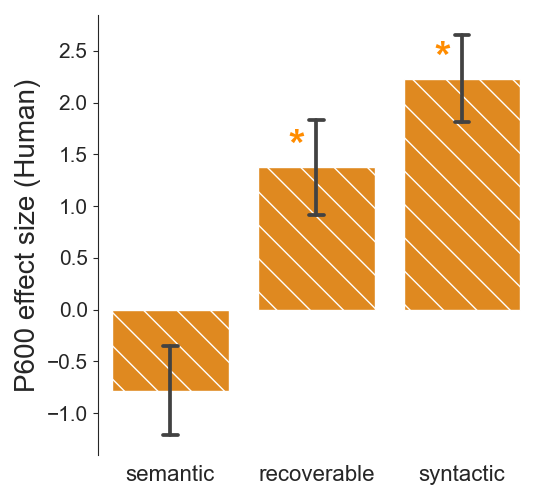}
         \caption{Ryskin-21 (Human)}
         \label{fig:ryskin_human}
    \end{subfigure}
    \caption{N400 and P600 amplitudes from model simulation and from human ERP experiments in Chow-16R (a-b), Chow-16S (c-d) and Ryskin-21 (e-f).}
    \label{fig:qualitative_results_part2}
\end{figure}

The last experiment (hereby \emph{Ryskin-21}) from~\citep{ryskin2021erp} has four conditions, one with a semantic violation (\emph{Semantic}), one with a syntactic violation (\emph{Syntactic}), one semantic critical condition (\emph{Recoverable}) with a semantic violation which could be attributed to noise, and a control sentence without any error (\emph{Control}). In the N400 time window, there is a significant N400 effect in \emph{Semantic} and \emph{Recoverable} conditions, where the N400 effect in \emph{Recoverable} condition is reduced. In the P600 time window, there is a significant P600 effect in \emph{Syntactic} and a smaller but significant P600 effect in \emph{Recoverable} condition (see Fig.~\ref{fig:ryskin_human}).

\begin{table*}[!htb]
\resizebox{\linewidth}{!}{
\centering
\begin{tabular}{lllll}\toprule
\textbf{Experiment} & \textbf{Condition} & \textbf{Context} & \textbf{Continuation} & \textbf{ERP} \\\midrule
\multirow{4}{*}{AD-98} & Syntactic & \multirow{4}{*}{\parbox{5cm}{The victims reported robbery}}  & police. & P600 \\
 & Semantic &  & to markets. & N400 \\
 & Double &  & markets. & Biphasic \\
 & Control & & to police. & NA \\\midrule
\multirow{3}{*}{Kim-05}  & Non-attractive & \parbox{5cm}{The dusty tabletop was} & devouring… & N400 \\\cmidrule{3-3}
 & Attractive & \multirow{2}{*}{\parbox{5cm}{The hearty meal was}} & devouring… & P600 \\
 & Control & & devoured… & NA \\\midrule
\multirow{4}{*}{Ito-16} & Semantic-related & \multirow{4}{*}{\parbox{7cm}{The student is going to the library to borrow a}} & page… & Reduced N400 \\
 & Form-related & & hook… & Biphasic \\
 & Unrelated & & sofa… & N400 \\
 & Control & & book… & NA \\\midrule
\multirow{3}{*}{Brothers-20S} & Unexpected & \multirow{3}{*}{\parbox{7cm}{James unlocked the}} & laptop... & N400 \\
 & Anomalous &  & gardener... & N400 \\
 & Control & & door... & NA \\\midrule
\multirow{3}{*}{Brothers-20L} & Unexpected & \multirow{3}{*}{\parbox{8cm}{He was thinking about what needed to be done on his way home. He finally arrived. James unlocked the}} & laptop... & N400 \\
 & Anomalous & & gardener... & Biphasic \\
 & Control & & door... & NA \\\midrule
\multirow{3}{*}{Brothers-20G} & Unexpected &  \multirow{3}{*}{\parbox{8cm}{The lifeguards received a report of sharks right near the beach. Their immediate concern was to prevent any incidents in the sea. Hence, they cautioned the}} & trainees & N400 \\
 & Anomalous & & drawer & Biphasic \\
 & Control & & swimmers & NA \\\midrule
\multirow{4}{*}{Chow-16R} & Reversal & The restaurant owner forgot which waitress the customer had & \multirow{2}{*}{served} & P600 \\
 & Reversal-Control & The restaurant owner forgot which customer the waitress had 
 &  & NA \\
 & Substitution1 & The superintendent overheard which neighbor the landlord had & \multirow{2}{*}{evicted} & Biphasic \\
 & Substitution1-Control & The superintendent overheard which exterminator the landlord had &  & NA \\\midrule
\multirow{4}{*}{Chow-16S} & Substitution2 & The tenant inquired which neighbor the landlord had & \multirow{4}{*}{evicted} & Biphasic \\
 & Substitution2-Control & The tenant inquired which exterminator the landlord had &  & NA \\
 & Swap & The exterminator inquired which neighbor the landlord had &  & N400 \\
 & Swap-Control & The neighbor inquired which exterminator the landlord had &  & NA \\\midrule
\multirow{4}{*}{Ryskin-21} & Semantic & \multirow{4}{*}{\parbox{8cm}{The storyteller could turn any incident into an amusing}} & hearse. & N400 \\
 & Syntactic & & anecdotes. & P600 \\
 & Recoverable & & antidote. & Biphasic \\
 & Control & & anecdote. & NA \\\bottomrule
\end{tabular}%
}
\caption{List of conditions, sample sentences and ERP patterns in dataset. The target word is the last word in the continuation.}
\label{tab:stimuli}
\end{table*}

\begin{figure}[!htb]
     \centering
     \captionsetup[subfigure]{justification=centering}
     \begin{subfigure}[b]{0.25\linewidth}
         \centering
         \includegraphics[width=\linewidth]{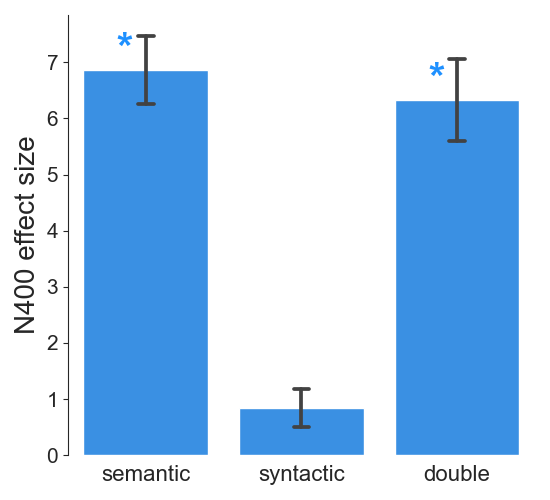}
         \includegraphics[width=\linewidth]{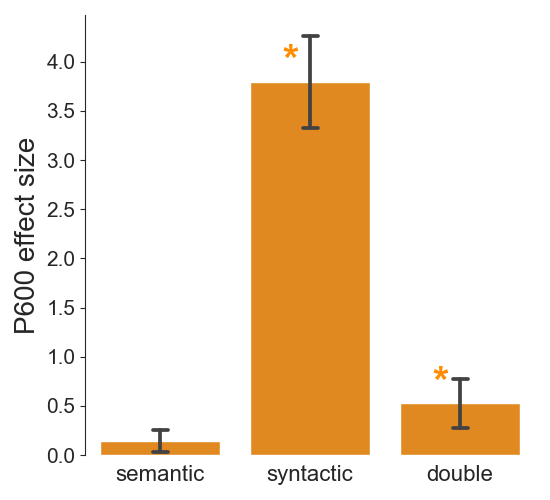}
         \caption{AD-98}
         \label{fig:syntax_gpt3}
     \end{subfigure}
   \hfill
     \begin{subfigure}[b]{0.15\linewidth}
         \centering
         \includegraphics[width=\linewidth]{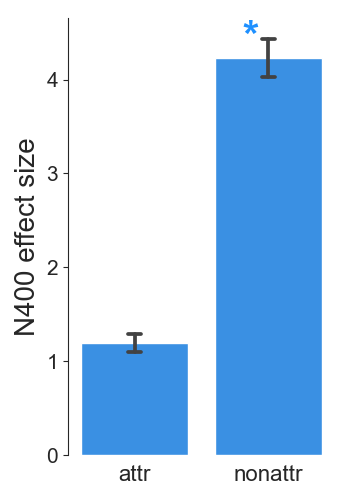}
         \includegraphics[width=\linewidth]{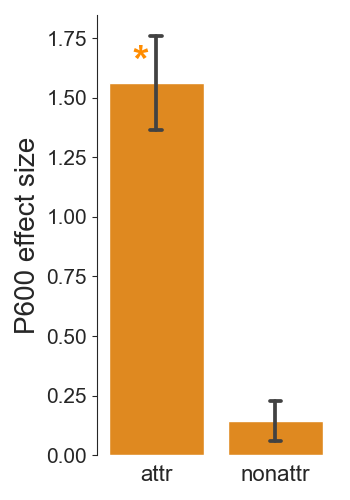}
         \caption{Kim-05}
         \label{fig:animacy_gpt3}
     \end{subfigure}
   \hfill
     \begin{subfigure}[b]{0.25\linewidth}
         \centering
         \includegraphics[width=\linewidth]{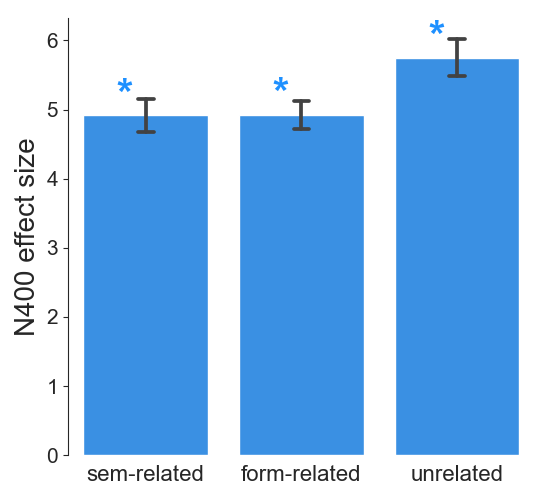}
         \includegraphics[width=\linewidth]{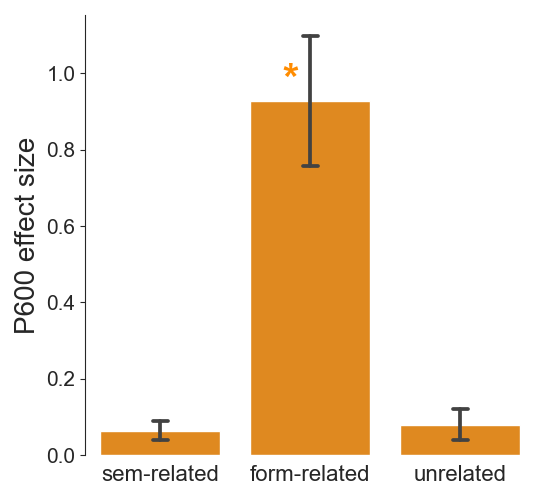}
         \caption{Ito-16}
         \label{fig:priming_gpt3}
     \end{subfigure}
     \hfill
     \begin{subfigure}[b]{0.25\linewidth}
         \centering
         \includegraphics[width=\linewidth]{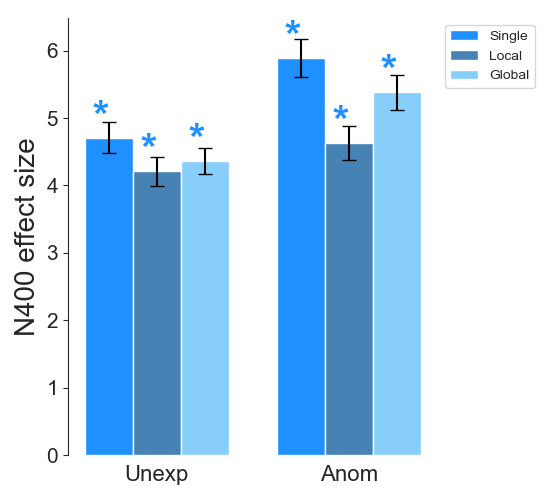}
         \includegraphics[width=\linewidth]{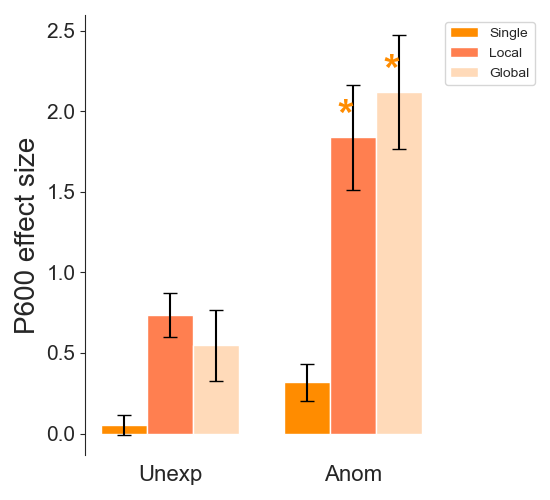}
         \caption{Brothers-20}
         \label{fig:brothers_gpt3}  
     \end{subfigure}
    \caption{N400 and P600 amplitudes from model simulation in AD-98 (a), Kim-05 (b), Ito-16 (c) and Federmeier-07 (d). Stars indicate a significant N400 or P600 in the original underlying studies as reported by the authors.}
    \label{fig:qualitative_results}
\end{figure}

\subsection{Qualitative Validation}

Figure~\ref{fig:qualitative_results} shows the simulated N400 and P600 effects across linguistic manipulations in datasets. Numerical tables of results are in SI~Section~2.

\paragraph{AD-98}  For AD-98~\citep{ainsworth1998dissociating} (see Fig.~\ref{fig:syntax_gpt3}), as expected, our model successfully simulated a larger N400 effect in conditions where there is a semantic violation (\emph{Semantic} and \emph{Double} conditions), and a larger P600 effect in \emph{Syntactic} and \emph{Double} conditions. The model seems to underestimate the size of the P600 effect in the \emph{Double} condition. This underestimation may be due to issues in the generation of shallow representation candidates. For example, given double-violation input \emph{The victim reported the robbery markets}, the candidate representations are forms like \emph{The victim reported the robbery masterminds} instead of the desired \emph{The victim reported the robbery to markets}. 

% Language model is inherently insensitive to syntactic errors so correcting the syntactic error doesn't give a big difference in B.

% RF: Since we still underestimate even with the right correction, the issue isn't in the way we're generating candidates. I think the underestimation is a general property of our model for these stimuli, unless there's some other correction we've missed which would give a big deep surprisal. We can discuss.
%RF: Can we say anything about the relative size of the effects in the experimental data?

\paragraph{Kim-05} Fig.~\ref{fig:animacy_gpt3} shows N400 and P600 effects in Kim-05~\citep{kim_independence_2005}. Consistent with human experimental results, the model simulated a greater N400 amplitude for \emph{Non-attractive} animacy violations, and a greater P600 response in the \emph{Attractive} condition, relative to the \emph{Control} condition. The success of the model relies on the fact that a larger proportion of sentences in \emph{Attractive} condition have plausible heuristic alternatives than in \emph{Non-attractive} condition.

\paragraph{Ito-16} In Ito-16~\citep{ito2016predicting} (see Fig.~\ref{fig:priming_gpt3}), the model correctly predicts an N400 effect in \emph{Form-related}, \emph{Semantic-related} and \emph{Unrelated} conditions, relative to \emph{Control} conditions, and a P600 effect in \emph{Form-related} condition. Importantly, consistent with human ERP patterns, the N400 effect is reduced for \emph{Semantic-related} and \emph{Form-related} conditions, with the N400 effect for \emph{Semantic-related} conditions being smallest. Our results provide distinct explanations for observed ERP effects in \emph{Semantic-related} and \emph{Form-related} conditions. In the \emph{Semantic-related} condition, the form-based distance term $d_\varphi(x,w)$ prevents semantically-related words from appearing in shallow representations. The GPT-2 probability of semantically-related words is smaller than the unrelated target, similar to the mechanism of N400 reduction induced by pre-activation of related semantic features. In contrast, in the \emph{Form-related} condition, the form-based distance term $d_\varphi(x,w)$ encourages shallow representations that are phonologically related to the input. The N400 reduction in the \emph{Form-related} condition is a result of assigning a more plausible representation to errors similar to the true prediction in the surface form. 

\paragraph{Brothers-20} In Brothers-20 \citep{brothers2020going} experiment (see Fig~\ref{fig:brothers_gpt3}), the model accurately predicts N400 effects to unexpected but plausible violations and to anomalous semantic violations, when following a single sentence with verbal constraint (\emph{Brothers-20S}). Moreover, the model successfully predicts P600 effects triggered by anomalous continuations when the violation was embedded within longer discourse contexts (\emph{Brothers-20L}) and when the precedding context was rich and globally constraining (\emph{Brothers-20G}). To model this dataset, we set the depth parameter $\lambda$ depending on the amount of discourse context, with less context corresponding to lower $\lambda$, reasoning that when the discourse provides rich semantic and contextual information, comprehenders rely more on their prior knowledge. 

\paragraph{Chow-16R} For Chow-16R~\citep{chow_bag--arguments_2016} experiment (see Fig.~\ref{fig:rev_gpt3} for model simulation and Fig.~\ref{fig:rev_human} for human results), the model correctly predicts a biphasic effect due to word substitution, and a P600 effect due to role reversals. Unlike the human data, the model simulates a non-negligible N400 for role reversals, which means that a significant proportion of role reversal sentences are interpreted literally. This effect appears to be a result of weakness in the GPT-2 language model: GPT-2 tends to underestimate the surprisal of role-reversal anomalies, even though human cloze probabilities differ greatly between role reversal and canonical control sentences (see SI~Section~2). 

\paragraph{Chow-16S} For Chow-16S~\citep{chow_bag--arguments_2016}, see Fig.~\ref{fig:swap_gpt3} for model simulation and Fig.~\ref{fig:swap_human} for human results. The model correctly predicts a biphasic effect for word substitution, and an N400 effect when words are swapped over a clause boundary.

\paragraph{Ryskin-21} For Ryskin-21~\citep{ryskin2021erp}  (see Fig.~\ref{fig:ryskin_gpt3} for model simulation and Fig.~\ref{fig:ryskin_human} for human results), the model successfully simulates an N400 for semantic violations, a P600 effect for syntactic violations, and a biphasic effect for recoverable semantic violations. 

\subsection*{Quantitative Validation}\label{sec:quantitative_validation}
A large body of existing work shows that N400 amplitudes can be predicted from language model surprisal \citep{frank2013word,michaelov2020well,michaelov2021different,michaelov2022more,merkx2020comparing,merkx2021human,frank2015erp,michaelov2024strong}. Our model suggests that this view is incomplete: LM surprisal will predict N400 amplitudes only when a shallow representation of the input suffices (that is, when shallow surprisal $A$ is equal to veridical surprisal), and the additional surprisal beyond what is captured in the shallow representations (the deep surprisal $B$) will surface as P600 amplitude.

We investigated this prediction in trial-level analysis on the last three experiments. 
We statistically confirmed the relationship between the empirical ERP amplitudes (N400 and P600) and our information-theoretic measures (shallow surprisal $A$ and deep surprisal $B$) in maximal linear mixed-effects models including by-subject and by-item intercepts and slopes \citep{barr2013random}. 
We use shallow surprisal $A$ as a single predictor to predict ERP amplitude in the N400 time window, 
and deep surprisal $B$ to predict ERP amplitude in the P600 time window. We additionally include two models where veridical surprisal is used to predict N400 and P600 amplitude as a comparison. 
The operationalization of N400 and P600 amplitude is based on analysis in original studies~\citep{chow_bag--arguments_2016,ryskin2021erp}. For Chow-16R and Chow-16S, we selected averaged ERP amplitudes between 300-500ms from six central electrodes as N400, and averaged ERP amplitudes between 700-900ms from six posterior electrodes as P600. For Ryskin-21, we selected eight central-parietal electrodes and two time windows --- 300-500ms and 600-800ms. The surprisal of target word is calculated with GPT-2. The results are summarized in Table~\ref{tab:quantitative_gpt3}. 

Overall, our proposed decomposition of surprisal provides a better fit of the overall ERP components than veridical surprisal alone. 
We find a significant main effect of shallow surprisal on N400 amplitude in all three experiments (Chow-16R: $t = -2.17, p < 0.05$; Chow-16S: $t = -2.60, p < .05$; Ryskin-21: $t = -7.32, p < .001$), and a significant main effect of deep surprisal on P600 amplitude in Chow-16R ($t = 1.66, p < 0.05$) and Ryskin-21 ($t = 2.79, p < .01$). We do not find evidence that the deep surprisal predicts P600 amplitude for Chow-16S; we believe that this is due to the fact that GPT-2 probabilities appear to be insensitive to swap errors, as discussed above. In comparison, we find no significant effect of veridical surprisal on P600. 

We note that our finding is not contradictory to the finding that LM surprisals are strong predictors of the N400 amplitude in naturalistic text~\citep{aurnhammer2019evaluating,michaelov2022so}:
in many cases, heuristic surprisal and veridical surprisal are similar, especially so in naturalistic text where errors are rare.

\begin{table*}[!htb]
\centering
%\resizebox{\linewidth}{!}{
\begin{tabular}{cccc@{\hspace{0.05cm}}cc}
\toprule
\multirow{2}{*}{Experiment} & \multicolumn{2}{c}{N400}& & \multicolumn{2}{c}{P600} \\ \cmidrule{2-3} \cmidrule{5-6}
 & shallow surprisal & veridical surprisal  && deep surprisal & veridical surprisal \\\midrule
Chow-16R & -0.56 (-2.17*) & -0.49 (-2.10*) && 1.66 (2.14*) & 0.35 (1.54) \\
Chow-16S & -0.60 (-2.60*) & -0.32 (-1.80) && 0.65 (1.55) & 0.22 (1.15)\\
Ryskin-21 & -0.82 (-7.32***) & -0.69 (-6.71***) && 1.49 (2.79**) & -0.05 (-0.54)\\\bottomrule
\end{tabular}%}
\caption{The effects of veridical surprisal $-\ln p(x \mid c)$, shallow surprisal $A$, and deep surprisal $B$ on ERP amplitudes in the N400 and P600 time range in the experiment from \citep{chow_bag--arguments_2016,ryskin2021erp}. Numbers are $\beta$ values ($t$-values). \emph{p} $<$ 0.05*, \emph{p} $<$ 0.01**, \emph{p} $<$ 0.001***. A negative effect in the N400 range indicates the standard N400 effect; a positive effect in the P600 range indicates the standard P600 effect.}
\label{tab:quantitative_gpt3}
\end{table*}
% \begin{figure}[!htb]
%     \centering
%     \includegraphics[width = .45\linewidth]{exp2_surprisal_sub1.png}
%     \includegraphics[width = .45\linewidth]{exp2_surprisal_rev.png}
%     \includegraphics[width = .45\linewidth]{exp2_surprisal_sub2.png}
%     \includegraphics[width = .45\linewidth]{exp2_surprisal_swap.png}
%     \caption{True surprisal of experimental and control items in sub-rev (top) and sub-swap (bottom) experiments.}
%     \label{fig:exp2_surprisal_distribution}
% \end{figure}

\subsection*{Novel Predictions}

Our theory laid out in Eqs.~\ref{eq:decomposition}--\ref{eq:proportions} predicts the following relationship between surprisal $S$ and ERP amplitudes for N400 and P600:
\begin{equation}
\label{eq:proportional-adding}
S = \frac{1}{\alpha} \mathrm{N400} + \frac{1}{\beta} \mathrm{P600},
\end{equation}
for positive $\alpha$ and $\beta$. In this section, we test this formula quantitatively in a trial-by-trial analysis of data from two experiments from~\citep{chow_bag--arguments_2016}.

Our theory makes the following predictions in a regression setting, which follow from re-arrangements of Eq.~\ref{eq:proportional-adding} (red color indicates a positive effect, blue color indicates a negative effect): there is a positive main effect of N400 and P600 on surprisal (``LM Surprisal'', Eq.~\ref{eq:lm_surprisal}); N400 is jointly predicted by P600 (with a negative sign) and surprisal (with a positive sign; ``LM N400'', Eq.~\ref{eq:lm_N400}), and P600 is predicted by N400 (with a negative sign) and surprisal (with a positive sign; ``LM P600''~\ref{eq:lm_P600}). We performed all of these regressions using the operationalization of ERPs and calculation of surprisal from the Quantitative Validation, including random intercepts and slopes corresponding to all main effects by subject and item. The results support our predictions (see Table~\ref{tab:exp2_sum_result}).

\begin{equation}
 \label{eq:lm_surprisal}
\mathrm{Surprisal} \sim \mathrm{\red{N400}} + \mathrm{\red{P600}}
\end{equation}
\begin{equation}
\label{eq:lm_N400}
\mathrm{N400} \sim \mathrm{\red{Surprisal}} - \mathrm{\blue{P600}}
\end{equation}
\begin{equation}
 \label{eq:lm_P600}
\mathrm{P600} \sim \mathrm{\red{Surprisal}}-\mathrm{\blue{N400}}  
\end{equation}

\begin{table*}[!htb]
\centering
\resizebox{.95\linewidth}{!}{%
\begin{tabular}{cccc@{\hspace{0.05cm}}ccc@{\hspace{0.05cm}}cc}
\toprule
\multirow{2}{*}{Exp} & \multicolumn{2}{c}{LM Surprisal}&& \multicolumn{2}{c}{LM N400}&& \multicolumn{2}{c}{LM P600}\\ \cmidrule{2-3} \cmidrule{5-6} \cmidrule{8-9}
 & \red{N400} & \red{P600}  && \red{Surprisal} & \blue{P600} && \red{Surprisal} & \blue{N400} \\\midrule
Chow-16R & \red{0.017} (2.65***) & \red{0.017} (3.35***) && \red{0.38} (4.01***) & \blue{-0.60} (-35.85***) && \red{0.39} (4.75***) & \blue{-0.69} (-24.69***)\\
Chow-16S & \red{0.005} (1.69***) & \red{0.005} (1.50***) && \red{0.27} (2.31***) & \blue{-0.54} (-22.58***) && \red{0.24} (2.03***) & \blue{-0.52} (-18.03***)\\
Ryskin-21 & \red{0.05} (6.19***) & \red{0.03} (5.05***) && \red{0.62} (6.93***) & \blue{-0.60} (-33.10***) && \red{0.47} (5.32***) & \blue{-0.75} (-33.07***)\\\bottomrule
\end{tabular}%
}
\caption{Experiment stimuli and statistical analysis of model predictions. Numbers are beta-values (t-values). \emph{p} $<$ 0.001***. }
\label{tab:exp2_sum_result}
\end{table*}

These results confirm previous findings of a trade-off between N400 and P600 amplitudes~\citep{aurnhammer2023single,kim2018individual,hodapp2021n400}. We show that the total amplitude of the N400 and P600 is proportional to surprisal, suggesting the mechanism for N400-P600 coupling might go beyond component overlap.

\section{Discussion}

We characterize language comprehension as an optimal trade-off between processing accuracy and depth. The idea of shallow vs. deep presentation is compatible with behavioral and neural studies of false perception, where comprehenders might perform shallow and incomplete processing of the perceptual input and fail to notice errors \citep{ferreira_good-enough_2002,ferreira_misinterpretation_2000,staub2024perceptual,muller2020negative,gibson_rational_2013}. The online shallow processing is sensitive to semantic and phonological overlap between prior prediction and presented input \citep{luce1998recognizing,failes2020blurring,failes2022using}. Beyond simple lexical cues, shallow representation might show sensitivity to complex discourse, pragmatic and structural cues \citep{nieuwland2008truth,xiang2015reversing,berkum1999semantic,giulianelli2024generalized,meister2024towards,giulianelli2023information}. The nature of the shallow representation and the relative importance of various linguistic and extra-linguistic cues in the formation of the representation should be explored in the future.

How is our model related to the broader neurolinguistic landscape? Because our model is formulated at an abstract information-theoretic level, it is compatible to various degrees with existing more algorithmic-level theories. Our use of contextualized word probability metrics is compatible with theoretical views of the N400 as the result of pre-activation of upcoming linguistic input as well as integration of the target word with previous context \citep{kutas2011thirty}. It is consistent with computational models in which N400 reflects the magnitude of a change to a shallow or heuristic representation of the input~\citep{rabovsky2018modelling,brouwer_neurocomputational_2017,brouwer_neurobehavioral_2021,fitz2019language,li2023heuristic,michaelov2020well,michalon2019meaning,eddine2024predictive}: these connections are detailed more formally in SI~Sections~3 and~4. Our model holds that N400 is associated with cognitive processes on a shallow representation that could differ from the true literal meaning of the input. This allows our model to provide good predictions of emerging experimental evidence on N400 blindness to semantic anomalies with a plausible alternative~\citep{chow_bag--arguments_2016,kim_independence_2005,kuperberg_separate_2016,van_herten_when_2006}.

We have hypothesized that P600 indexes the effort of representation shift over time, in line with multiple strands of theoretical frameworks~\citep{kim_independence_2005,van_herten_erp_2005,leckey2020p3b,van2012prediction,kuperberg2020tale,brothers2020going}. We advances previous theories by formalizing the evolution from shallow to deep representation of input as a function of processing depth, and that P600 reflects cost of shifting from shallow to deep representation. Our model provides a unified theory to P600 elicited by syntactic violation \citep{hagoort_syntactic_1993}, semantic violation that bears orthographic or phonological similarity with the predictable target \citep{kim_independence_2005,van_herten_erp_2005,van_herten_when_2006,ryskin2021erp}, and semantic violation without orthographic relation with the predictable target in a richer discourse \citep{kuperberg2020tale,delong2020comprehending,brothers2020going,van2012prediction}. Previous computational work has explored various quantifications of P600 component, ranging from cosine similarity between contexts~\citep{li2023heuristic,brouwer_neurocomputational_2017}, to surprisal and entropy reduction~\citep{frank2015erp}, to probability ratio between shallow and literal representation~\citep{ryskin2021erp}, to the size of a gradient update to a language model \citep{fitz2019language}. We evaluate the ability of the gradient-based model to account for the experimental data here in SI~Section~3, finding that its performance is not as accurate as our shallow processing-based approach. Our model provides a principled integration of both the N400 and P600 signals: they reflect successively deeper linguistic processing, unfolding according to a comprehension process whose dynamics are dictated by a trade-off between accuracy and information processing.

\section*{Materials and Methods}
\paragraph{Derivation of shallow representation policy} 
The representation policy is the distribution $p(w \mid x, c)$ on representations $w$ given inputs $x$ in context $c$ that solves the following optimization problem, minimizing a distortion metric $d(w, x)$ subject to a constraint on how much information is extracted about the input $x$:
\begin{align}
\label{eq:opt}
&\mathop{\text{minimize }}_{p(\cdot \mid x, c)} \mathop\mathbb{E}\left[d(w,cx)\right] \\
\nonumber
&\text{subject to }\mathrm{D}_\text{KL}[p(w \mid x, c) \| p_0(w \mid c)] = C, 
\end{align}
where $\mathrm{D}_{\text{KL}}$ is the KL divergence from the prior $p_0(w \mid c)$ to the shallow representation policy $p(w \mid x, c)$, and $C$ is the amount of KL divergence permitted from $p_0$ to $p$. The constrained optimization problem of Eq.~\ref{eq:opt} can be converted to the problem of minimizing the unconstrained objective function
\begin{equation}
J(p(\cdot \mid x, c)) = \mathop\mathbb{E}\left[d(w,cx)\right] - \lambda D_\text{KL}[p(w \mid x, c) \| p_0(w\mid c)],
\end{equation}
where $\lambda$ is a Lagrange multiplier. The minimizing distribution has the form
\begin{equation}
p_\lambda(w \mid x, c) = \frac{1}{Z_\lambda(x, c)} p_0(w \mid c) e^{-\lambda d(w,cx)},
\end{equation}
%where $p(w)$ is the marginal distribution on representations, which we interpret as a prior over representations, 
where $Z_\lambda(x, c)$ is a normalizing constant and $\lambda$ is chosen to satisfy the constraint $\mathrm{D}_\text{KL}[p(w \mid x) \| p_0(w)]=C$.

\paragraph{Shallow surprisal} Given the optimal representation policy for depth parameter $\lambda$, the shallow surprisal comes out to
\begin{equation}
\mathrm{D}_{\text{KL}}[p_\lambda(w \mid x, c) \| p_0(w \mid c)] = - \ln Z_\lambda(x, c) - \lambda \mathop\mathbb{E}\left[ d(w, cx) \right].
\end{equation}

\paragraph{Distortion metric} For the form-based distance $d_\varphi(w,x)$ we use orthographic Damerau-Levenshtein edit distance, which augments Levenshtein edit distance to include character swaps as a possible edit operation. In addition to character-level edits, we allow word-level swaps where the cost of swapping two words within the memory span ($\leq 7$) is dependent on the number of words in between. For the semantic distance $d_\sigma(w,x)$, we use cosine distance calculated using GPT-2 embeddings \citep{radford2019language}. For the prior $p_0(w \mid c)$, we use the GPT-2 language model.

\paragraph{Computing Averages} 
In order to compute Eq.~\ref{eq:decomposition}, we need a way to average over all the possible shallow word strings $w$ given the input word strings $x$. This averaging is difficult because there are in principle an infinite number of possible shallow representations $w$ for any given input $x$. Therefore, we compute averages approximately, limiting the support of $w$ to only a subset of likely candidates. We generate candidates by prompting GPT-3 \citep[specifically \texttt{text-davinci-002},][]{brown2020language}. Results using a variety of other candidate generation methods are found in SI~Section~4. For each experimental stimulus, we input prompts with instructions and four examples on how to correct sentences to recover its intended meaning \footnote{Prompt: The final word in each of the following sentences is wrong: someone typed the wrong word. Please type in a different word, the one that was probably intended. \textbf{Input}: The hearty meal was devouring. \textbf{Correction}: The hearty meal was devoured. \textbf{Input}: The hearty meal was devoured. \textbf{Correction}: The hearty meal was devoured. \textbf{Input}: Mary went to the library to borrow a hook. \textbf{Correction}: Mary went to the library to borrow a book. \textbf{Input}: Mary went to the library to borrow a plant. \textbf{Correction}: Mary went to the library to borrow a plant. \textbf{Input}: \textit{Experimental Sentence} \textbf{Correction}:}, and ask GPT-3 to generate one best correction candidate. We generate ten candidates for each sentence at temperature 0.95, and then reweight them using Eq.~\ref{eq:heuristic-policy}. Results using alternative methods of generating candidate representations are similar, and shown in SI Section 4.

\paragraph{Free parameters} The scalar free parameters are $\lambda$, representing processing depth, and $\gamma$, representing the relative importance of form-based and meaning-based factors. Following the theory developed above, the $\gamma$ parameter should be constant, a fixed property of the human language comprehension system, whereas the processing depth parameter $\lambda$ plausibly varies across experiments. Many factors plausibly influence $\lambda$, including the proportion of plausible or implausible sentences in the stimuli, the nature of task demands, presentation latency and the type of contextual information~\citep{gunter1997syntax,hahne1999electrophysiological,zwaan1998situation,chow_wait_2018,brothers2020going,kuperberg2020tale}.

The numerical value of the parameter $\gamma=8$ should not be interpreted na\"ively to mean that semantic information is more important than form-based information in the formation of shallow representations. The two metrics $d_\phi$ and $d_\sigma$ as we quantify them are on different scales, and the parameter $\gamma$ is a conversion factor between them.

To fit values of $\gamma$ and $\lambda$, we first set $\gamma$ to be 1 and explored the effect of $\lambda$ with a grid search from 0.1 to 2, with a step size of 0.1, and with two marginal conditions ($\lambda = 0$ and $\lambda = 10$). For each experiment, we selected the $\lambda$ based on visual inspection of the simulated ERP pattern. Table~\ref{tab:lambda} shows a list of $\lambda$ values used in simulating the presented results. Next, having set $\lambda$ to the values, we performed a grid search of $\gamma$ from 0 to 10 with a step size of 1. We did another round of visual inspection of the simulated ERP patterns and selected $\gamma = 8$ for all experiments.

\begin{table}[!htb]
     \centering
     %\resizebox{.95\linewidth}{!}{
     \begin{tabular}{cccccccc}\toprule
       & AD-98 & Kim-05 & Ito-16 & Brothers-20 & Chow-16R & Chow-16S &Ryskin-21\\\midrule
     $\lambda$  & 2 & 0.8 & 1.5 & 1.5(S)/0.8 (L\&G) & 0.8 & 0.6 & 1\\\bottomrule
     \end{tabular}
     %}
     \caption{Values of processing depth parameter $\lambda$ across experiments.}
     \label{tab:lambda}
\end{table}

% \showmatmethods{} % Display the Materials and Methods section

% \acknow{We thank the Computational Psycholinguistics Lab at MIT, the UCI Quantitative Language Collective, and \dots for helpful comments.}

% \showacknow{} % Display the acknowledgments section

\bibliographystyle{apacite}

\setlength{\bibleftmargin}{.125in}
\setlength{\bibindent}{-\bibleftmargin}
\bibliography{main}

\newpage
\appendix
\section*{\LARGE \textbf{Supplementary Information}}

\section{Representation Policy as Noisy Channel Inference}
\label{sec:si-noisy-channel-math}

Our representation policy has the form
\begin{equation}
\label{eq:heuristic-policy}
p(w \mid x) \propto p_0(w) \exp{-\lambda d(w, x)},
\end{equation}
where $d(w,x)$ is a distortion metric between inputs $x$ and representations $w$. (We have suppressed conditioning on the context $c$ for simplicity of notation.) This is equivalent to noisy-channel inference under the following conditions. Suppose that a speaker's intended utterance $w$ is corrupted into a received input $x$ according to a noise distribution, $p_N(x \mid w)$. If the distortion $d(w,x)$ is equal to the log-likelihood of $x$ given $w$ under this model, that is
\begin{equation}
\label{eq:distortion-noise}
d(w, x) = -\ln p_N(x \mid w),
\end{equation}
then substituting Eq.~\ref{eq:distortion-noise} into Eq.~\ref{eq:heuristic-policy} gives a representation policy
\begin{align}
p(w \mid x) &\propto p_0(w) \exp{\lambda \ln p_N(x \mid w)} \\
&= p_0(w) p_N(x \mid w)^\lambda,
\end{align}
which is identical when $\lambda=1$ to Bayes' rule for prior $p_0$ and likelihood $p_N$, thus describing rational inference over intended utterances $w$ given noisy inputs $x$.

\section{Numerical Results}
Table~\ref{tab:si-surprisal} shows average numerical values of the simulation results. LM surprisal is the surprisal calculated from GPT-2, and veridical surprise is the surprisal of the veridical input re-normalized among all possible candidate representations. 
\begin{table*}[!htb]
    \centering
    \resizebox{\linewidth}{!}{
    \begin{tabular}{cccccc}
    \toprule
      Experiment & Condition & shallow surprisal & deep surprisal & Veridical surprisal & LM surprisal\\\midrule
\multirow{4}{*}{AD-98} & Syntactic &  1.97 & 3.81 & 5.78 & 13.52\\
 & Semantic & 7.99 & 0.17 & 8.15 & 13.41\\
 & Double & 7.45 & 0.55 & 8.00 & 14.74\\
 & Control & 1.12 & 0.02 & 1.14 & 7.90\\\midrule
\multirow{3}{*}{Kim-05} & Attractive & 1.52 & 1.74 & 3.27 & 8.72\\
 & Non-attractive &  4.56 & 0.33 & 4.89 & 9.42\\
 & Control & 0.33 & 0.18 & 0.51 & 5.76\\\midrule
\multirow{4}{*}{Ito-16} & Semantic-related & 6.18 & 0.06 & 6.24 & 8.38 \\
 & Form-related & 6.19 & 0.93 & 7.11 & 9.22 \\
 & Unrelated & 7.01 & 0.08 & 7.09 & 9.26\\
 & Control & 1.26 & 0.00 & 1.26 & 3.50 \\\midrule
 \multirow{3}{*}{Brothers-20S} & Unexpected & 4.81 & 0.08 & 4.89 & 9.27 \\
 & Anomalous & 5.99 & 0.34 & 6.34 & 10.76 \\
 & Control & 0.10 & 0.02 & 0.13 & 4.69 \\\midrule
  \multirow{3}{*}{Brothers-20L} & Unexpected &  4.44 & 1.05 & 5.49 & 9.17 \\
 & Anomalous & 4.86 & 2.14 & 7.01 & 10.66 \\
 & Control & 0.23 & 0.30 & 0.54 & 4.14 \\\midrule
 \multirow{3}{*}{Brothers-20G} & Unexpected & 4.49 & 1.17 & 5.66 & 8.69 \\
 & Anomalous & 5.51 & 2.74 & 8.25 & 11.05 \\
 & Control & 0.12 & 0.62 & 0.75 & 3.80 \\\midrule
 \multirow{4}{*}{Chow-16R} & Reversal & 1.48 & 0.83 & 2.31 & 5.49\\
 & Reversal-Control & 0.47 & 0.13 & 0.60 & 5.50 \\
 & Substitution1 & 2.06 & 0.16 & 2.23 & 5.35 \\
 & Substitution1-Control & 0.66 & 0.07 & 0.74 & 4.87\\\midrule
 \multirow{4}{*}{Chow-16S} & Substitution2 & 1.52 & 0.56 & 2.07 & 5.59 \\
 & Substitution2-Control & 0.70 & 0.21 & 0.91 & 5.12 \\
 & Swap & 1.52 & 0.35 & 1.86 & 5.56\\
 & Swap-Control & 0.54 & 0.31 & 0.85 & 5.16\\\midrule
 \multirow{4}{*}{Ryskin-21} & Semantic & 4.74     & 0.21 & 4.95 & 8.74\\
 & Syntactic & 1.05 & 1.79 & 2.84 & 7.53\\
 & Recoverable & 2.63 & 1.39 & 4.03 & 8.49 \\
 & Control & 0.53 & 0.079 & 0.61 & 5.20\\\bottomrule
    \end{tabular}}
    \caption{Average veridical surprisal, shallow surprisal and deep surprisal across experiments}
    \label{tab:si-surprisal}
\end{table*}

\section{Comparison with other models}

We discuss the relationship between our model and other existing computational models predicting the N400 and/or P600 signals. For the Error Propagation model, we provide an empirical comparison against our model.

\paragraph{N400 from Language Models}
Many exploratory analyses of linguistic ERP signals have used measures estimated from large-scale neural network models trained on natural languages\citep{frank2013word,frank2015erp,michaelov2020well,merkx2021human,michaelov2021different,michaelov2022more}. N400 is typically operationalized with metrics that capture the relationship between context and target (cosine similarity or language model surprisal), whereas P600 is under-explored. The success of neural networks may reflect the influence of a kind of shallow representation, because neural network representations are driven by notions of plausibility and statistical regularities rather than a strictly faithful representation of the literal meaning of the sentence~\citep{mccoy2019berts,ettinger2020bert,min2020syntactic}. Our model is in agreement with the existing models that equate language model surprisal to the N400 signal inasmuch as language model surprisals reflect shallow representations---when language model surprisals differ from shallow representations, that is when deep processing is required, then we predict a reduced N400 and a P600. In line with \citep{michaelov2024mathematical}'s finding that the relationship between LM surprisal and N400 amplitude may be sub-logarithmic, our model predicts N400 amplitudes that are generally lower than would be expected from surprisal, because our N400 measure is upper-bounded by the surprisal of the current input: 
\begin{equation}
\mathrm{D}_\text{KL}\left[p_{\lambda}(w \mid x, c) \| p_0(w \mid c)\right] \le -\ln p_0(x \mid c).
\end{equation}

\paragraph{Information value} \citet{giulianelli2024generalized} proposed a framework for expressing surprisal measures weighted by semantic distance to alternatives, called Information Value. They show that different distance metrics within this framework are predictive of different EEG signals. Our theory differs in that we have a constant distortion metric among forms, which is a combination of phonological and semantic distance; different EEG signals correspond to different depths (weights $\lambda$) rather than different metrics. 

Our measure of shallow surprisal is similar to but mathematically distinct from Information Value. Information Value generally takes the form
\begin{equation}
\mathrm{IV} = \mathop\mathbb{E}_{p_0}\left[d(x, w)\right],
\end{equation}
where $w$ is an expected word, $x$ is an observed word, $p_0$ is a language model (distribution on expected words $w$ given some context), and $\displaystyle\mathop\mathbb{E}_{p_0}$ indicates an expectation over expected words $w$ from the language model (dependence on context $c$ is suppressed in the notation in this section). In contrast, our shallow surprise can be written
\begin{align}
\mathrm{D}_{\text{KL}}\left[p_\lambda(w \mid x) \| p_0(w)\right] &= \mathop\mathbb{E}_{p_{\lambda}}\left[\ln \frac{p_\lambda(w \mid x)}{p_0(w)}\right] \\
&= \mathop\mathbb{E}_{p_\lambda}\left[\ln\frac{\frac{1}{Z_\lambda(x)} \cancel{p_0(w)} e^{-\lambda d(x,w) }}{\cancel{p_0(w)}} \right] \\
&= -\ln Z_\lambda (x) - \lambda \mathop\mathbb{E}_{p_\lambda}\left[d(x, w)\right] \\
&= -\ln \mathop\mathbb{E}_{p_0}\left[e^{-\lambda d(x,w)} \right] - \lambda \mathop\mathbb{E}_{p_\lambda}\left[d(x,w)\right].
\end{align}

Like Information Value, shallow surprisal also involves weighted expectations over alternatives, but rather than an expectation of distance with respect to the prior distribution $p_0$, we have two terms: (1) the log of the expectation with respect to $p_0$ of the exponentially-decaying similarity of alternatives and (2) the expectation of distance with respect to the \emph{representation policy} $p_\lambda$. The similarity of these theories suggests that there may be a deeper connection between them.

\paragraph{Sentence Gestalt} 

We show that the Sentence Gestalt model \citep{rabovsky2018modelling} of the N400 response may be interpreted in the framework of our model, for a particular form of the shallow representations and representation policy. 

In the Sentence Gestalt (SG) model, the N400 reflects an update to internal activations in a neural network that probabilistically predicts sentence meaning~\citep{rabovsky2018modelling,rabovsky2020quasi,rabovsky2020change,lindborg2021meaning,lindborg2022n400,lindborg2023semantic,hodapp2021n400}. This measure reflects cognitive principles that comprehenders predict a representation of meaning, and N400 amplitude is induced by the change in the prediction given new upcoming stimulus. Specifically, let $\phi(x)$ be a $K$-dimensional vector representation of the meaning of input string $x$. The SG model holds that N400 response to current input $x_t$ is given by the magnitude of the change to the representation $\phi$ in response to the new input:
\begin{equation}
\label{eq:sg-update}
\mathrm{N400} \propto \sum_{k=1}^K | \phi_k(x_{\le t}) - \phi_k(x_{<t}) |.
\end{equation}

Though the SG model does not have an explicit theory involving shallow representations, the vector-valued semantic representation can be interpreted as a shallow representation that cares solely about event plausibility. Our model is formally related to the SG model under this view. In our model, the N400 is proportional to the magnitude of the update from the prior distribution on representations $p_0(w \mid x_{<t})$ to the shallow representation policy $p_\lambda(w \mid x_{\le t})$ given new input $x_t$, measured as KL divergence:
\begin{equation}
\label{eq:kl-update}
\mathrm{N400} \propto \mathrm{D}_{\text{KL}}\left[p_\lambda(w \mid x_{\le t}) \| p_0(w \mid x_{<t})\right].
\end{equation}
In the main text, we considered the shallow representations $w$ to range over strings, with the representation policy taking a form determined by optimally trading off distortion and KL divergence. Instead, we now consider the possibility that shallow representations $w$ are $K$-dimensional vectors, and that the shallow representation policy $p(w \mid x)$ is a Normal distribution centered around a real-valued vector $\phi(x)$ with standard deviation $\sigma^2$:
\begin{equation}
w \mid x \sim \mathcal{N}(\phi(x), \sigma^2).
\end{equation}
Then the KL divergence in Eq.~\ref{eq:kl-update} comes out to the magnitude of the change in the representations $\phi$:
\begin{equation}
\mathrm{N400} \propto \frac{1}{2\sigma^2} \sum_{k=1}^K \left|\phi_k(x_{\le t}) - \phi_k(x_{<t})  \right|^2,
\end{equation}
which is formally similar to the SG update in Eq.~\ref{eq:sg-update}, except that the SG update involves absolute differences, whereas the KL divergence involves squared difference. 

\paragraph{Retrieval-Integration model} The Retrieval Integration model \citep{brouwer_neurocomputational_2017} proposes a single-stream framework for N400 and P600. The N400 reflects difficulty of lexical retrieval in context, and the P600 reflects difficulty of lexical integration in context. ERP responses are simulated as change of internal activation patterns in retrieval and integration layers of a neural network model \citep{brouwer_neurocomputational_2017}, or as change of reconstructed utterances (surprisal) \citep{brouwer_neurobehavioral_2021}. The shallow representation is present implicitly, as the frequency distribution in the training data: the training data have more samples with a canonical thematic-role assignment (e.g. waitress serves customer) than noncanonical assignment (e.g. customer serves waitress). This setup reflects the prior world knowledge and provides a basis for the shallow representation. After changing the relative frequency of agent-patient pairs in the training data, the model fails to simulate the desired ERP pattern~\citep{ettinger_relating_2018}. 
 
\paragraph{Noisy channel model}
Noisy channel accounts of N400 and P600 have held that, given an input $x$, and comprehender forms a representation $w$ of what the speaker likely meant, by Bayesian noisy-channel inference. Then the N400 is proportional to the surprisal of the inferred representation $w$, and P600 is proportional to the probability ratio of $x$ and $w$ \citep{ryskin2021erp}, or to some other measure of divergence between $x$ and $w$ \citep{li2023heuristic}. 

As shown in Appendix Section~\ref{sec:si-noisy-channel-math}, our shallow representation policy is identical to noisy-channel inference for a certain choice of the distortion metric. Thus, our model is broadly compatible with the noisy-channel approach to linguistic ERP signals. We differ from existing noisy-channel accounts in the exact formulas used to predict N400 and P600 from these representations, while sharing intuitive motivations with these work. In \citep{ryskin2021erp}, N400 and P600 magnitudes are directly linked to the posterior probability ratio between a single inferred shallow representation and a single literal representation. In \citep{li2023heuristic}, the noisy channel selection determines the characteristics of the shallow representations. N400 amplitude is linked to the conditional probability of target given shallow context, whereas P600 amplitude is linearly correlated with cosine similarity of vector representation between shallow and veridical representations. Our model shares the intuition with these two models by assuming that N400 and P600 amplitudes correlated with the probabilistic nature of shallow representations, but advances previous models by positing an entire probabilistic \emph{distribution} of shallow representation. 

Our model differs from noisy-channel accounts on the cognitive explanation of shallow representations. In noisy-channel accounts, shallow representations are in fact rationally-inferred representations based on a veridical perception of the input, whereas in our account, they are shallow but `good-enough' representations based on incomplete processing of information about the input \citep{ferreira2002good,ferreira2007good}. Our model thus naturally predicts the ordering of N400 and P600: the two signals reflect the dynamics of on-going processing of linguistic input, proceeding from shallow to deep representations \citep{trueswell1993verb,frazier1982making,macdonald_lexical_1994,gouvea2010linguistic,hoeks_seeing_2004,van_gompel_unrestricted_2000,bever1970cognitive,frazier1978sausage,tabor2004effects}. In noisy-channel models, by contrast, there is no particular explanation for why the P600 signal follows the N400 signal, and in fact the ordering is problematic if one assumes that rational inference follows veridical perception of the input.

%\subsection*{Interim Summary} The idea of shallow representations gives an intuitive explanation of N400 and P600 patterns. This idea has been explicitly formalized (e.g. Noisy-Channel model) or implicitly implemented via characteristics of vectorized representation (e.g. large language models). In these computational models, N400 is approximated with some computational metrics that quantify the relationship between vectorized heuristic representation and target word. This is consistent with cognitive theories where N400 reflects context-based pre-activation \citep{kutas2000electrophysiology,delong2005probabilistic,wicha2003potato,van2005anticipating,cheimariou2019lexical,szewczyk2018n400} or semantic integration of input \citep{hagoort2009semantic} or a combinatory process of both \citep{van2009neuropragmatics,calloway2017integrative, nieuwland2020dissociable}. P600 metrics involve interaction between heuristic and literal representation, consistent with the view that P600 amplitude reflects syntactic reanalysis \citep{kim_independence_2005}, effort of integrating structural information \citep{kaan2000p600,bornkessel2008alternative} or effortful error monitoring process~\citep{kolk2003structure,van_herten_erp_2005,van_herten_when_2006,van2011monitoring}.

\paragraph{Error Propagation}
In \citet{fitz2019language}, N400 and P600 ERPs during sentence comprehension reflect adaptation and learning to the statistical regularities in the linguistic environment. Specifically, P600 reflects the size the model update, which would be quantified as the size of prediction error propagation in neural network. 

\begin{figure}[!htb]
     \centering
     \captionsetup[subfigure]{justification=centering}
     \begin{subfigure}[b]{0.3\linewidth}
         \centering
         \includegraphics[width=\linewidth]{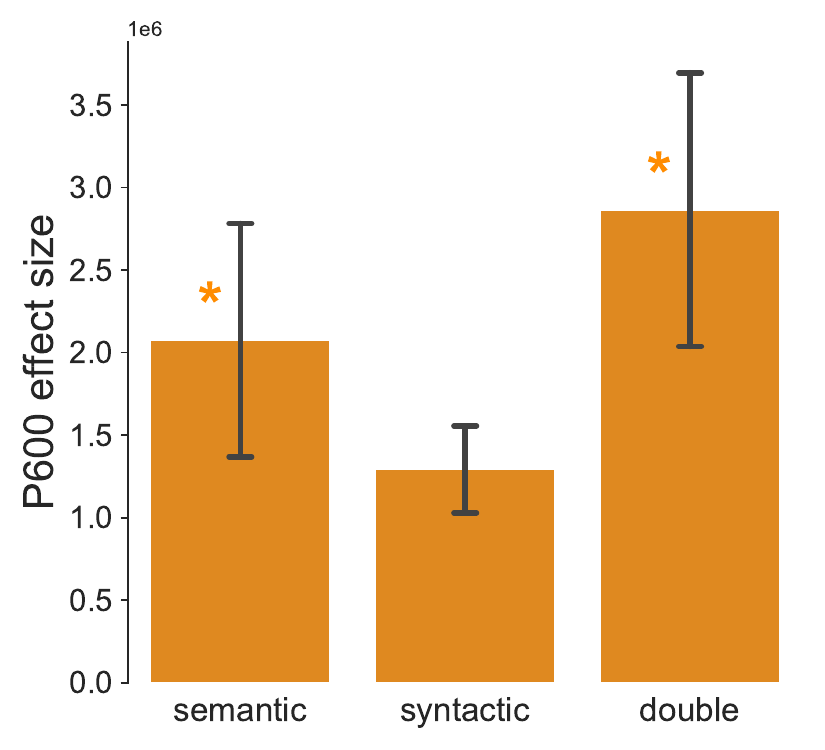}
         \caption{AD-98}
         \label{fig:syntax_gradient}
     \end{subfigure}
     %\hfill
     \begin{subfigure}[b]{0.17\linewidth}
         \centering
         \includegraphics[width=\linewidth]{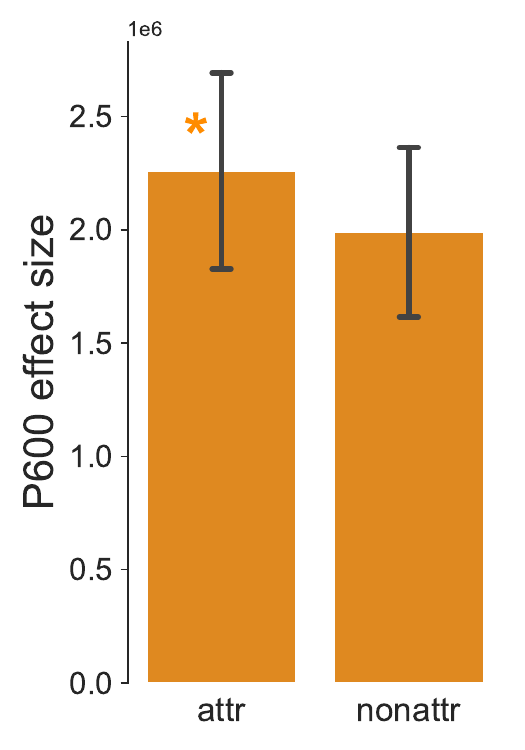}
         \caption{Kim-05}
         \label{fig:animacy_gradient}
     \end{subfigure}
     %\hfill
     \begin{subfigure}[b]{0.3\linewidth}
         \centering
         \includegraphics[width=\linewidth]{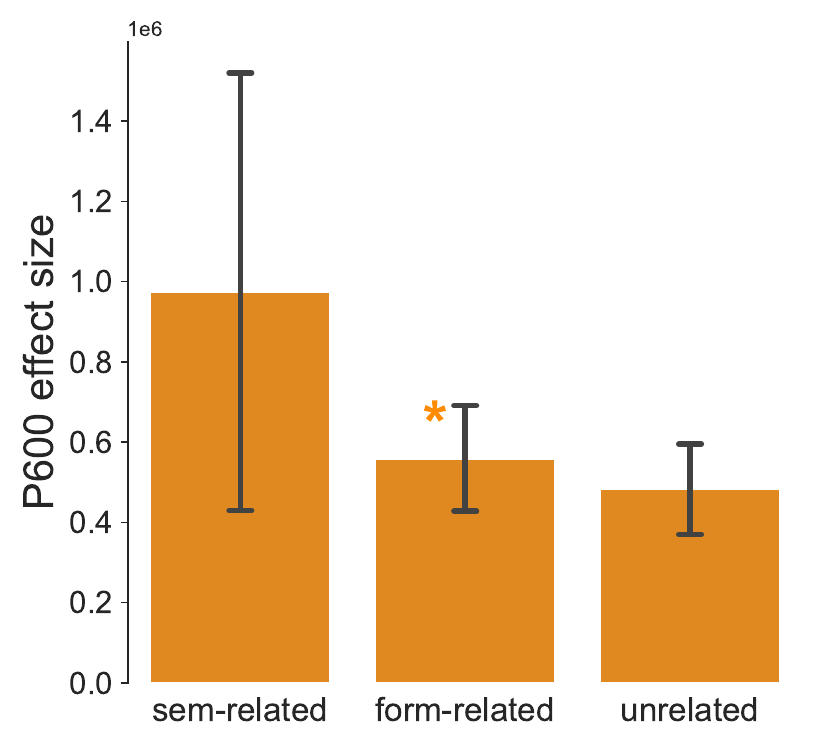}
         \caption{Ito-16}
         \label{fig:priming_gradient}
     \end{subfigure}
     \begin{subfigure}[b]{0.3\linewidth}
         \centering
         \includegraphics[width=\linewidth]{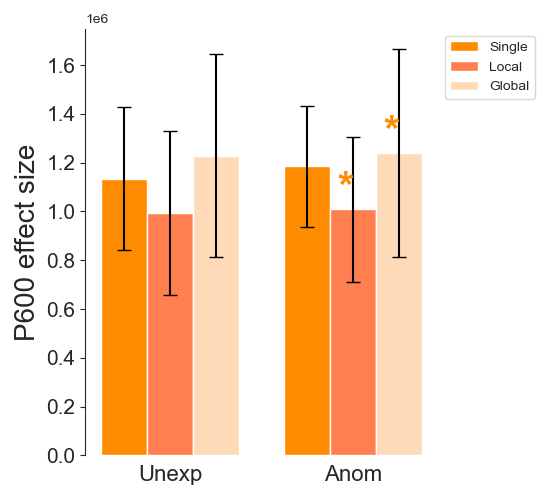}
         \caption{Brothers-20}
         \label{fig:brothers_gradient}  
     \end{subfigure}
     % \captionsetup[subfigure]{justification=centering}
     %\hspace{1.5cm}
    \begin{subfigure}[b]{0.17\linewidth}
         \centering
         \includegraphics[width=\linewidth]{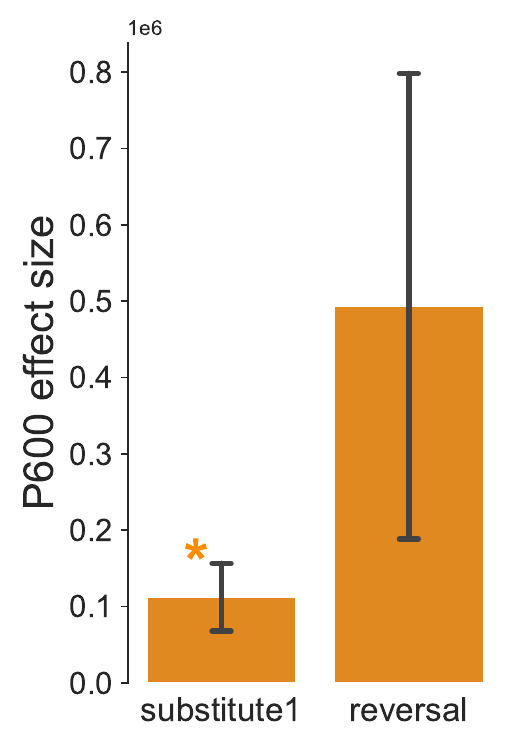}
         \caption{Chow-16R}
         \label{fig:rev_gradient}
     \end{subfigure}
     %\hspace{0.5cm}
    \begin{subfigure}[b]{0.17\linewidth}
         \centering
         \includegraphics[width=\linewidth]{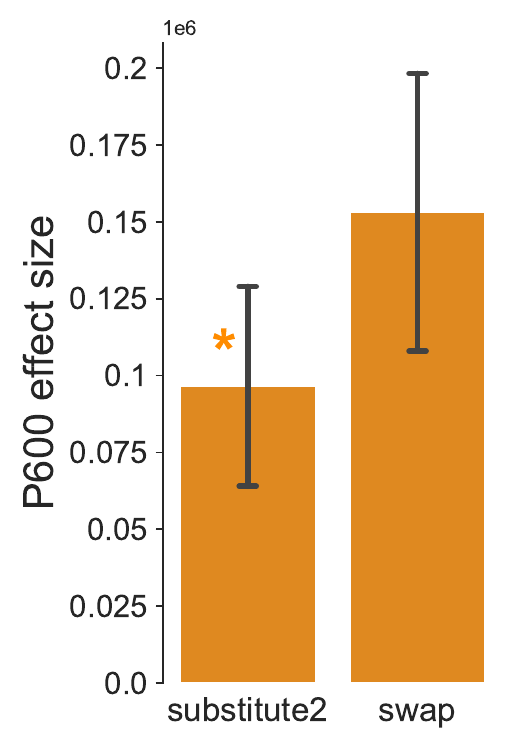}
         \caption{Chow-16S}
         \label{fig:swap_gradient}
     \end{subfigure}  
     % \hspace{0.5cm}
     \begin{subfigure}[b]{0.3\linewidth}
         \centering
         \includegraphics[width=\linewidth]{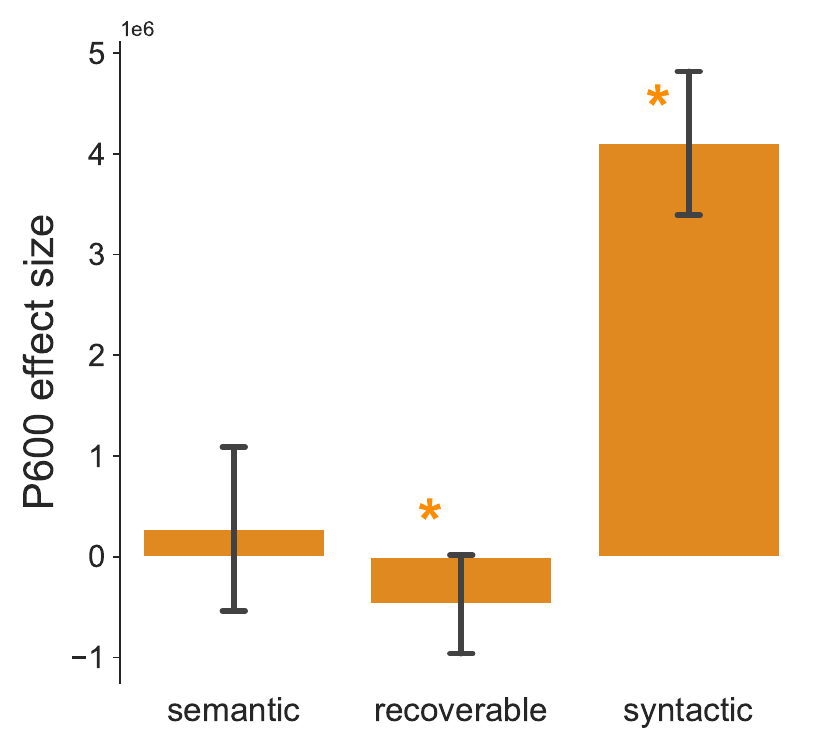}
         \caption{Ryskin-21}
         \label{fig:ryskin_gradient}
     \end{subfigure}
    \caption{P600 amplitudes simulated with Error Propagation model using GPT-2.}
    \label{fig:gradient_results}
\end{figure}

We evaluate this model against the same dataset of experimental results that we used in the main text. We use the off-the-shelf GPT-2 (small) language model, quantifying P600 as the summed absolute value of gradients induced by the target word across all units in the GPT-2 internal layers. Figure~\ref{fig:gradient_results} shows the simulated P600 effect size across the experiments used in the current study. The model qualitatively reproduces the relative order of the P600 effect in AD-98 and Kim-05, but fails to simulate the ERP patterns in other experiments. Furthermore, this theory does not straightforwardly predict the relationship between N400, P600, and surprisal predicted by our theory and validated in the main text.

\section{Simulations using other methods for candidate generation}

For the distribution on shallow words given input words, we provide two additional implementations, which differ in their strengths. The first method is to manually construct shallow candidates based on original experimental setups. The second method generate shallow candidates by sampling from semantic and phonological competitors, and other highly probable continuations of the prior context. 

\paragraph{Control Counterpart}
Though GPT-3 prompting provides an objective and effective way to generate candidate corrections, it is not transparent on the generation process. To mitigate this problem, we systematically constructed shallow candidate words as the control sentences in the original experiments. For \emph{The hearty meal was devouring...} from \emph{Attractive} condition in Kim-05, the shallow representation distribution include two candidates: the original sentence \emph{The hearty meal was devouring...} and the control counterpart \emph{The hearty meal was devoured...}. We calculated the shallow representation using the method described in Section 2, and set the value of $\gamma$ to be the same 8 and performed a grid-search for $\lambda$. Table~\ref{tab:lambda-control} the list of $\lambda$ values used in simulating the presented results. 

\begin{table*}[!htb]
     \centering
     %\resizebox{.95\linewidth}{!}{
     \begin{tabular}{cccccccc}\toprule
       & AD-98 & Kim-05 & Ito-16 & Brothers-20 & Chow-16R & Chow-16S & Ryskin-21\\\midrule
     $\lambda$  & 1.2 & 0.5 & 1.5 & 1.2(S)/1.0(L\&G) & 0.8 & 0.6 & 1\\\bottomrule
     \end{tabular}
     %}
     \caption{Control Counterpart: Values of processing depth parameter $\lambda$ across experiments}
     \label{tab:lambda-control}
\end{table*}

\begin{figure}[!htb]
     \centering
     \captionsetup[subfigure]{justification=centering}
     \begin{subfigure}[b]{0.3\linewidth}
         \centering
         \includegraphics[width=\linewidth]{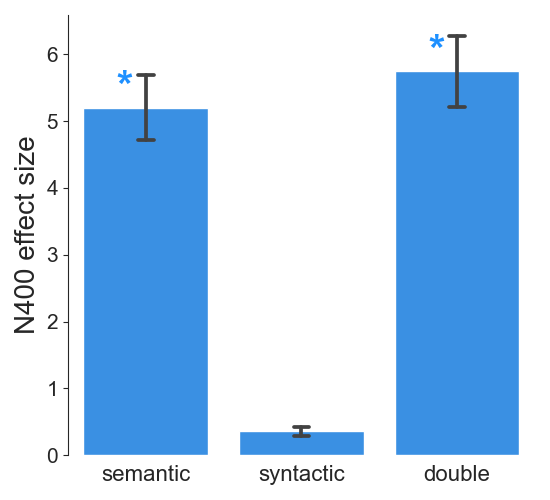}
         \includegraphics[width=\linewidth]{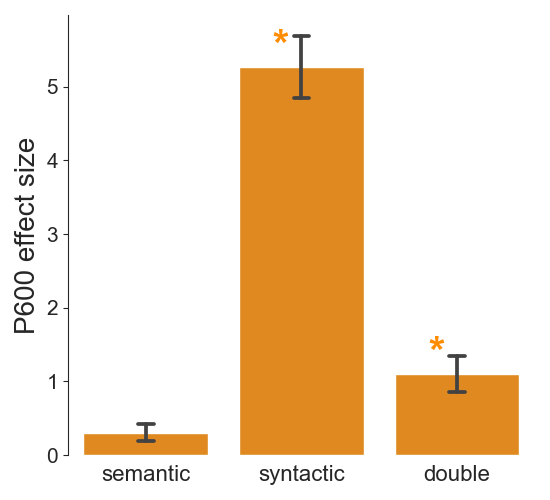}
         \caption{AD-98}
         \label{fig:syntax_counterpart}
     \end{subfigure}
     %\hfill
     \begin{subfigure}[b]{0.17\linewidth}
         \centering
         \includegraphics[width=\linewidth]{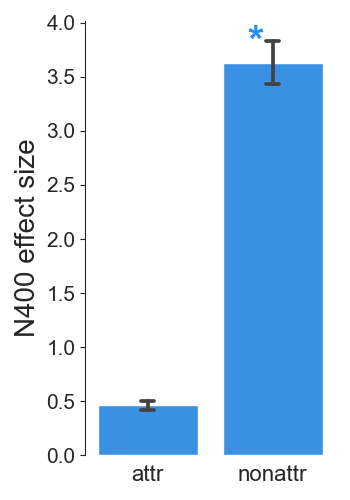}
         \includegraphics[width=\linewidth]{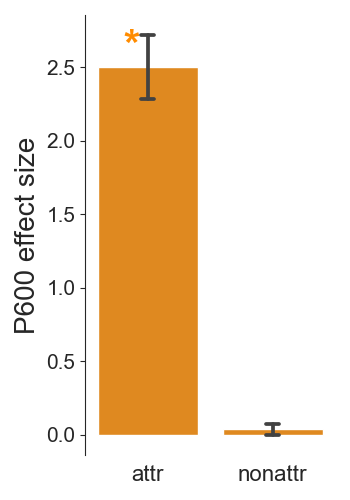}
         \caption{Kim-05}
         \label{fig:animacy_counterpart}
     \end{subfigure}
     %\hfill
     \begin{subfigure}[b]{0.3\linewidth}
         \centering
         \includegraphics[width=\linewidth]{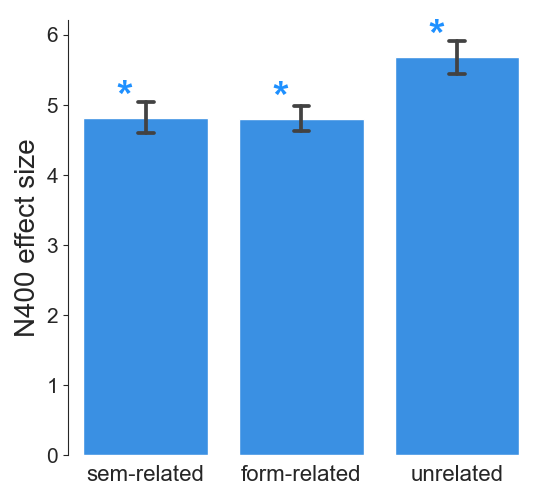}
         \includegraphics[width=\linewidth]{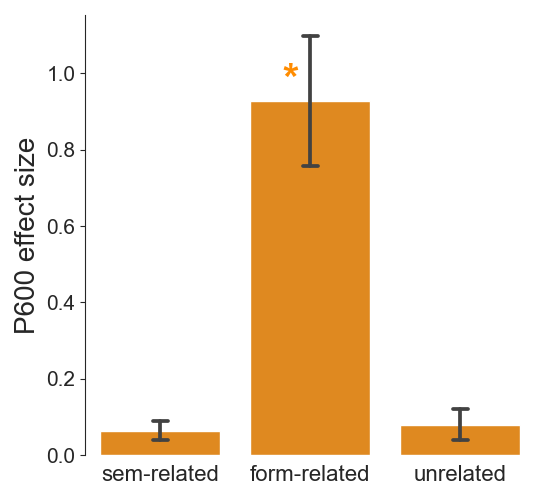}
         \caption{Ito-16}
         \label{fig:priming_counterpart}
     \end{subfigure}
     \begin{subfigure}[b]{0.3\linewidth}
         \centering
         \includegraphics[width=\linewidth]{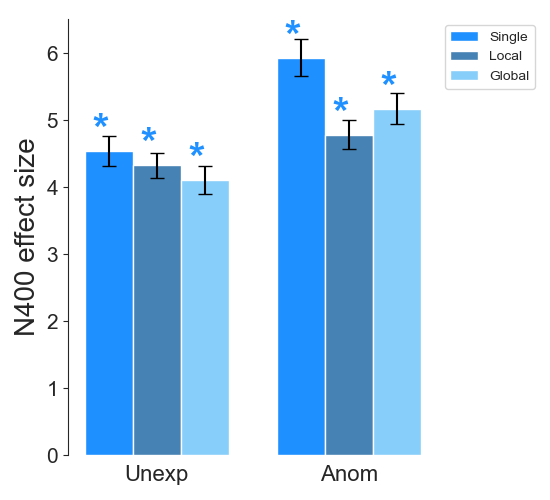}
         \includegraphics[width=\linewidth]{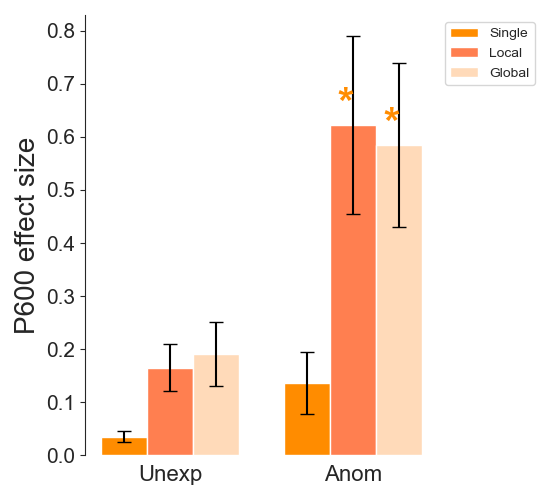}
         \caption{Brothers-20}
         \label{fig:brothers_counterpart}  
     \end{subfigure}
      %\vfill
     % \captionsetup[subfigure]{justification=centering}
     %\hspace{1.5cm}
    \begin{subfigure}[b]{0.17\linewidth}
         \centering
         \includegraphics[width=\linewidth]{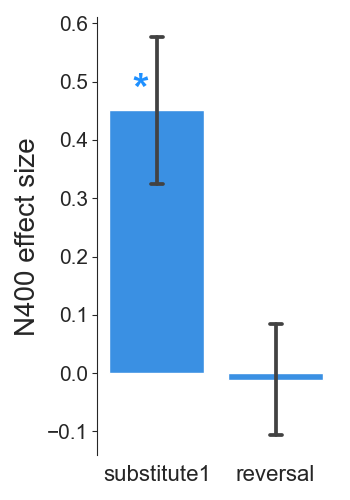}
         \includegraphics[width=\linewidth]{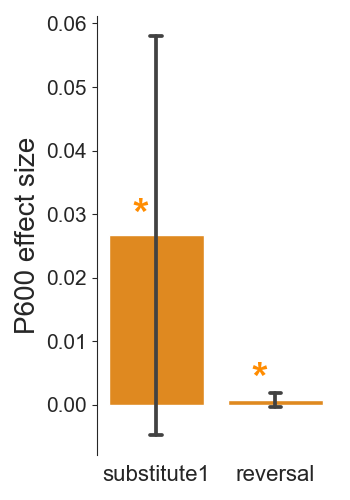}
         \caption{Chow-16R}
         \label{fig:rev_counterpart}
     \end{subfigure}
     %\hspace{0.5cm}
    \begin{subfigure}[b]{0.17\linewidth}
         \centering
         \includegraphics[width=\linewidth]{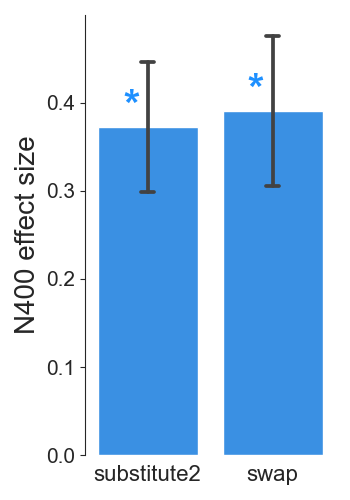}
         \includegraphics[width=\linewidth]{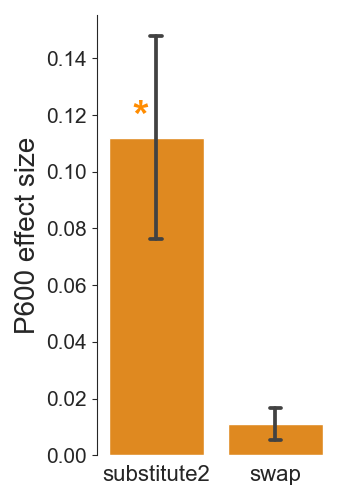}
         \caption{Chow-16S}
         \label{fig:swap_counterpart}
     \end{subfigure}  
     % \hspace{0.5cm}
     \begin{subfigure}[b]{0.3\linewidth}
         \centering
         \includegraphics[width=\linewidth]{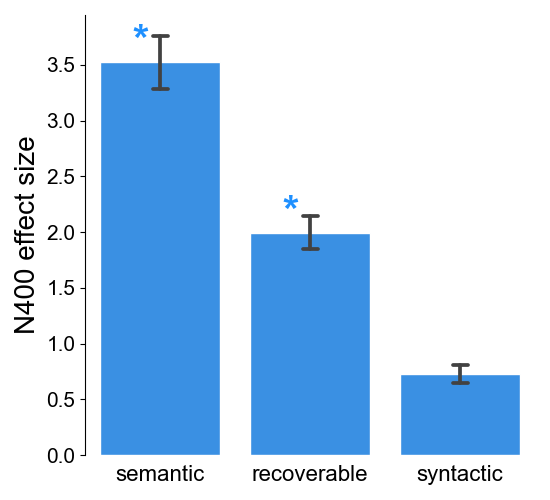}
         \includegraphics[width=\linewidth]{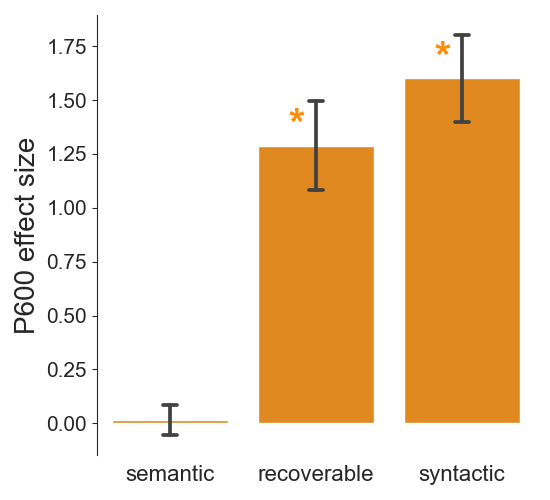}
         \caption{Ryskin-21}
         \label{fig:ryskin_counterpart}
     \end{subfigure}
    \caption{N400 and P600 amplitudes simulated with the counterpart sentences in the other conditions as candidate representations.}
    \label{fig:manual_results}
\end{figure}

The simulation results are presented in Fig.~\ref{fig:manual_results}. The model quantitatively simulated ERP patterns in most experiments, however, the simulation fails to replicate P600 effect to role-reversal anomalies (Chow-16R). This is because GPT-2 assigns similar probability to canonical targets and role-reversed violations (see Table \ref{tab:si-surprisal}).

We also conducted trial-level analysis using linear mixed-effects model. Table \ref{tab:quantitative_control} shows the statistical results. Our shallow surprisal and deep surprisal significantly predict N400 ($t = -5.69***$) and P600 ($t = 2.99**$) respectively in Ryskin-21, however, none of the predictors significantly predict the ERP patterns in Chow-16R and Chow-16S. 

\begin{table*}[!htb]
\centering
%\resizebox{0.75\linewidth}{!}{
\begin{tabular}{cccc@{\hspace{0.05cm}}cc}
\toprule
\multirow{2}{*}{Experiment} & \multicolumn{2}{c}{N400}& & \multicolumn{2}{c}{P600} \\ \cmidrule{2-3} \cmidrule{5-6}
 & shallow surprisal & veridical surprisal  && deep surprisal & veridical surprisal \\\midrule
Chow-16R & -1.05 (-1.53) & -1.05 (-1.53) && -10.06 (-1.39) & -0.12 (-0.18) \\
Chow-16S & -0.78 (-1.70) & -0.59 (-1.53) && 1.09 (0.87) & -0.05 (-0.12)\\
Ryskin-21 & -0.85 (-5.69***) & -0.63 (-5.05***) && 2.46 (2.99**) & 0.04 (0.35)\\\bottomrule
\end{tabular}%}
\caption{Control Counterpart: The effects of veridical surprisal $-\ln p(x \mid c)$, shallow surprisal $A$, and deep surprisal $B$ on ERP amplitudes in the N400 and P600 time range in the experiment from \citep{chow_bag--arguments_2016,ryskin2021erp}. Numbers are $\beta$ values ($t$-values). \emph{p} $<$ 0.05*, \emph{p} $<$ 0.01**, \emph{p} $<$ 0.001***. A negative effect in the N400 range indicates the standard N400 effect; a positive effect in the P600 range indicates the standard P600 effect.}
\label{tab:quantitative_control}
\end{table*}

\paragraph{Multiple Sampler}
In Multiple Sampler generation, we construct candidate alternatives by pooling potential candidates from three sources: semantic competitors, phonological competitors and top predictions based on prior context. For semantic and phonological competitors, we sampled 100 closest phonological competitors from the 60,000 most frequent English words in SUBTLEXus \citep{brysbaert2009moving}. The phonological distance is approximated as edit distance between words. We also select 100 semantic competitors by choosing words with closest cosine distance of non-contextualized GloVe embeddings \citep{pennington2014glove}. We sample contextual congruent candidates by selecting another 100 words with highest probability given the prior context in GPT-2. After getting candidate shallow target words from these three sources, we replace the true target word with candidate target in the original sentence to get the candidate sentence. For each presented sentence, there are 300 alternatives. The posterior probability distribution of representation candidates is calculated using representation policy in Equation \ref{eq:heuristic-policy}. We chose the value of $\gamma = 8$ and performed a grid-search of $\lambda$ values similar as the procedure described in other generation methods. Table~\ref{tab:lambda-competitor} the list of $\lambda$ values used in simulating the results using Multiple Sampler. 

\begin{table*}[!htb]
     \centering
     %\resizebox{.95\linewidth}{!}{
     \begin{tabular}{cccccccc}\toprule
       & AD-98 & Kim-05 & Ito-16 & Brothers-20 & Chow-16R & Chow-16S & Ryskin-21\\\midrule
     $\lambda$  & 1.5 & 1.2 & 1.5 & 1.5(S)/1.0(L\&G)  & 0.8 & 1.0 & 1.2\\\bottomrule
     \end{tabular}
     %}
     \caption{Multiple Sampler: Values of processing depth parameter $\lambda$ across experiments}
     \label{tab:lambda-competitor}
\end{table*}

\begin{figure}[!htb]
     \centering
     \captionsetup[subfigure]{justification=centering}
     \begin{subfigure}[b]{0.3\linewidth}
         \centering
         \includegraphics[width=\linewidth]{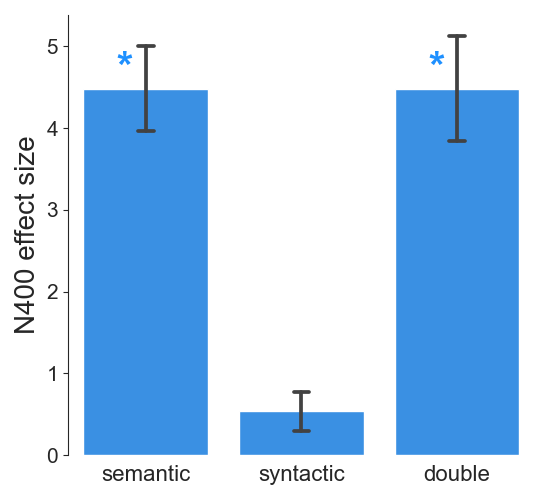}
         \includegraphics[width=\linewidth]{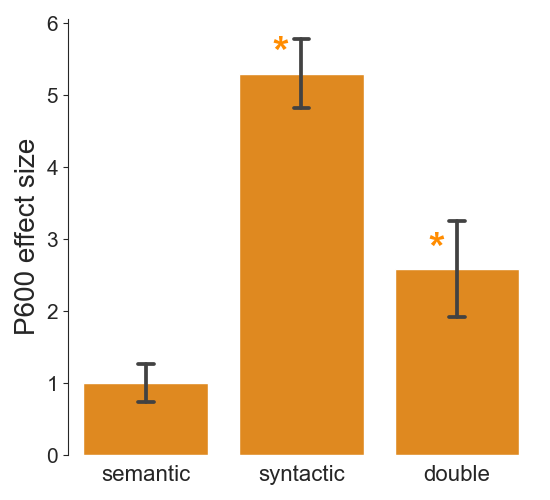}
         \caption{AD-98}
         \label{fig:syntax_competitor}
     \end{subfigure}
     %\hfill
     \begin{subfigure}[b]{0.17\linewidth}
         \centering
         \includegraphics[width=\linewidth]{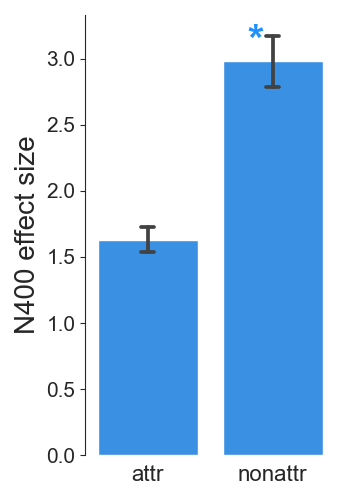}
         \includegraphics[width=\linewidth]{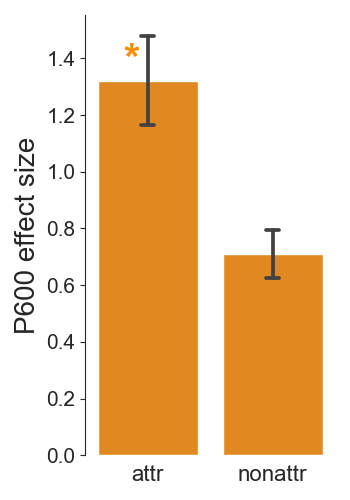}
         \caption{Kim-05}
         \label{fig:animacy_competitor}
     \end{subfigure}
     %\hfill
     \begin{subfigure}[b]{0.3\linewidth}
         \centering
         \includegraphics[width=\linewidth]{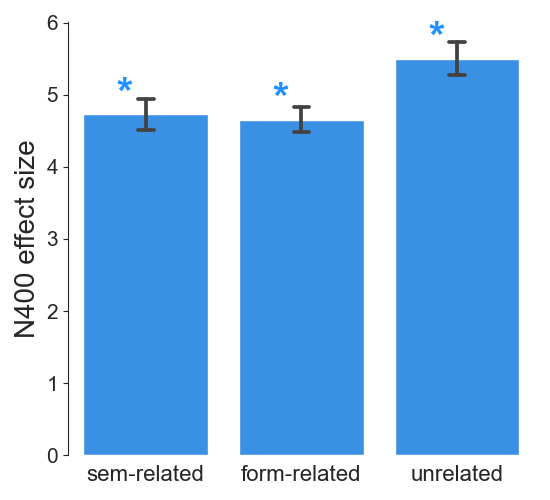}
         \includegraphics[width=\linewidth]{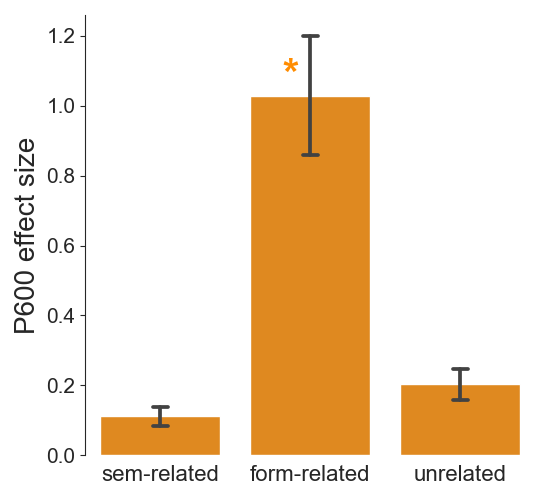}
         \caption{Ito-16}
         \label{fig:priming_competitor}
     \end{subfigure}
    \begin{subfigure}[b]{0.3\linewidth}
         \centering
         \includegraphics[width=\linewidth]{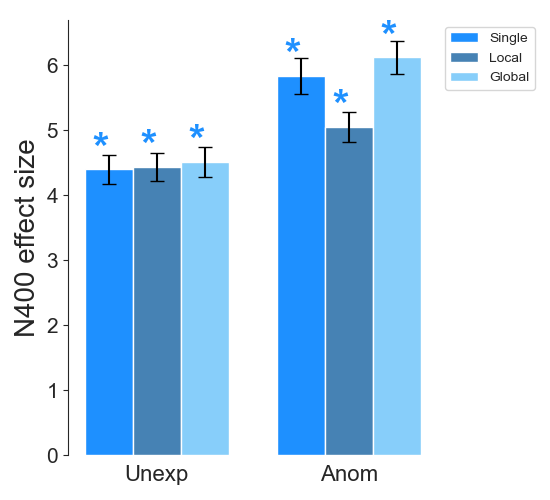}
         \includegraphics[width=\linewidth]{figures/brothers_competitor_n400.png}
         \caption{Brothers-20}
         \label{fig:brothers_competitor}  
     \end{subfigure}
     % \captionsetup[subfigure]{justification=centering}
     %\hspace{1.5cm}
    \begin{subfigure}[b]{0.17\linewidth}
         \centering
         \includegraphics[width=\linewidth]{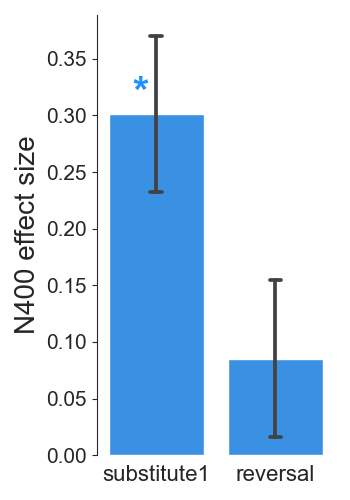}
         \includegraphics[width=\linewidth]{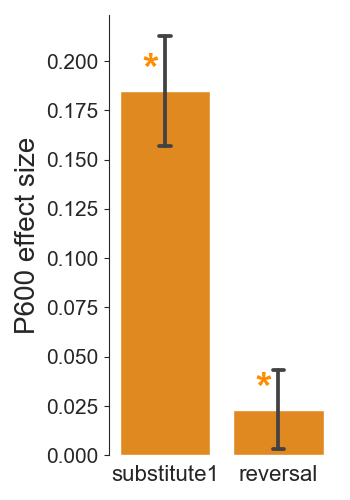}
         \caption{Chow-16R}
         \label{fig:rev_competitor}
     \end{subfigure}
     %\hspace{0.5cm}
    \begin{subfigure}[b]{0.17\linewidth}
         \centering
         \includegraphics[width=\linewidth]{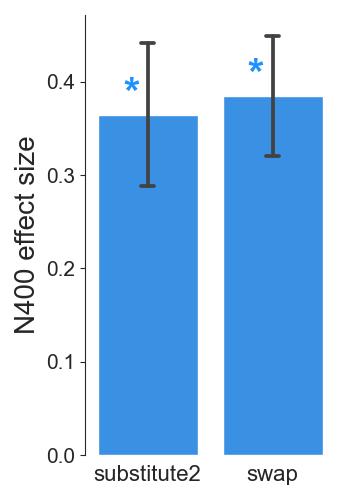}
         \includegraphics[width=\linewidth]{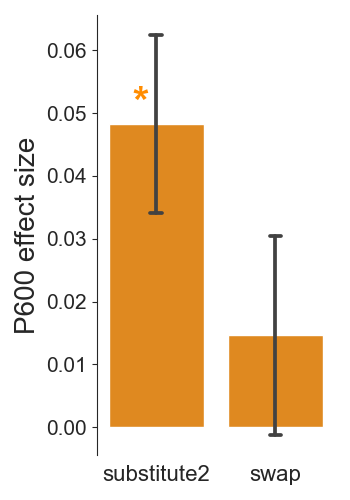}
         \caption{Chow-16S}
         \label{fig:swap_competitor}
     \end{subfigure}  
     % \hspace{0.5cm}
     \begin{subfigure}[b]{0.3\linewidth}
         \centering
         \includegraphics[width=\linewidth]{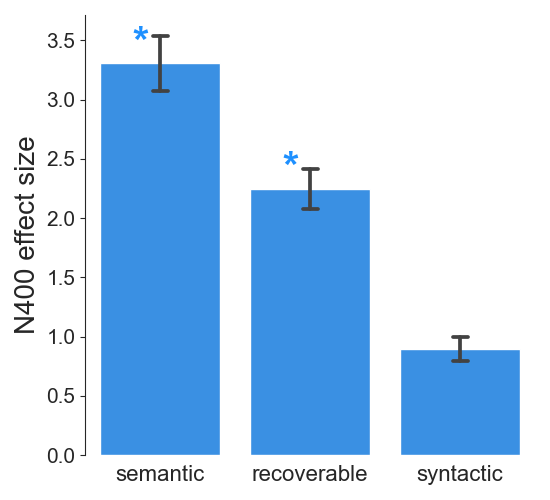}
         \includegraphics[width=\linewidth]{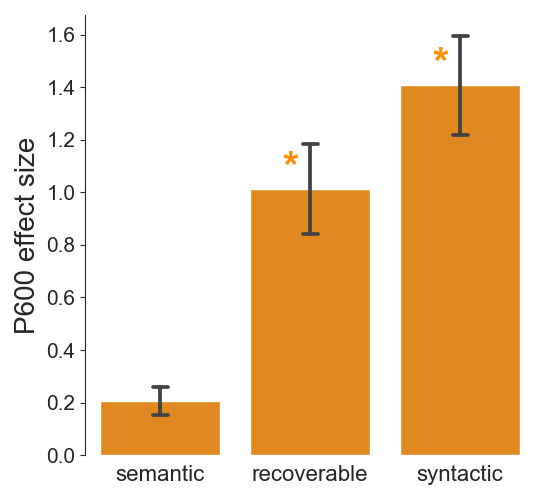}
         \caption{Ryskin-21}
         \label{fig:ryskin_competitor}
     \end{subfigure}
    \caption{N400 and P600 amplitudes simulated with alternative representations sampled from phonological and semantic competitors and top language model predictions from prior context.}
    \label{fig:competitor_results}
\end{figure}

Figure \ref{fig:competitor_results} shows a qualitative patterns of simulated N400 and P600 using Multiple Sampler. The model qualitatively reproduce the ERP patterns across most experiments, but show weaker difference between experimental and control conditions in Chow-16R and Chow-16S. This might be because the shallow candidate only considers revision at the target word position, and therefore unable to explain linguistic manipulations involving word transportation. 

Table \ref{tab:quantitative_competitor} shows the quantitative results on the last three experiments. As expected, the model fails to track ERP patterns for Chow-16R and Chow-16S. In Ryskin-21, the shallow surprisal is a significant predictor of N400 ($t = -6.08***$), but the deep surprisals does not significantly predict P600 amplitude ($t = 1.52$).

\begin{table*}[!htb]
\centering
%\resizebox{0.75\linewidth}{!}{
\begin{tabular}{cccc@{\hspace{0.05cm}}cc}
\toprule
\multirow{2}{*}{Experiment} & \multicolumn{2}{c}{N400}& & \multicolumn{2}{c}{P600} \\ \cmidrule{2-3} \cmidrule{5-6}
 & shallow surprisal & veridical surprisal  && deep surprisal & veridical surprisal \\\midrule
Chow-16R & -0.47 (-0.79) & -0.98 (-2.54*) && -0.37 (0.88) & 0.82 (1.85) \\
Chow-16S & -0.31 (-0.86) & -0.29 (-1.02) && 0.18 (0.24) & 0.12 (0.45)\\
Ryskin-21 & -0.83 (-6.08***) & -0.68 (-5.69***) && 0.78 (1.52) & 0.03 (0.21)\\\bottomrule
\end{tabular}%}
\caption{Multiple Sampler: The effects of veridical surprisal $-\ln p(x \mid c)$, shallow surprisal $A$, and deep surprisal $B$ on ERP amplitudes in the N400 and P600 time range in the experiment from \citep{chow_bag--arguments_2016,ryskin2021erp}. Numbers are $\beta$ values ($t$-values). \emph{p} $<$ 0.05*, \emph{p} $<$ 0.01**, \emph{p} $<$ 0.001***. A negative effect in the N400 range indicates the standard N400 effect; a positive effect in the P600 range indicates the standard P600 effect.}
\label{tab:quantitative_competitor}
\end{table*}
\end{document}